%% file: main.tex
\pdfoutput=1
\documentclass[preprint,12pt]{elsarticle}
\usepackage{amsmath,amssymb}
\usepackage{algorithmic}
\usepackage{algorithm}
\usepackage{array}
\usepackage[caption=false,font=normalsize,labelfont=sf,textfont=sf]{subfig}
\usepackage{textcomp}
\usepackage{stfloats}
\usepackage{url}
\usepackage{verbatim}
\usepackage{graphicx}
\usepackage{float}
\usepackage{tikz}
\usepackage{makecell}
\usepackage{multirow}
\usepackage{hyperref}
\usepackage{tabularx}
\usepackage{xcolor}
\definecolor{olive}{rgb}{0.33, 0.42, 0.18}
\usepackage{color}
\usepackage{pdflscape}
\usepackage[normalem]{ulem}

\begin{document}

\begin{frontmatter}

\title{
    A Variational Latent-Space Framework for Uncertainty-Aware Spectral Image Emulation
\tnoteref{preprint-note}
}

\tnotetext[preprint-note]{This article has been submitted to \textit{Science of Remote Sensing}.}

\author[1]{Chedly Ben Azizi\corref{cor1}}
\ead{chedly.ben-azizi@univ-littoral.fr}
\author[1]{Claire Guilloteau}
\ead{claire.guilloteau@univ-littoral.fr}
\author[1]{Gilles Roussel}
\ead{gilles.roussel@univ-littoral.fr}
\author[1]{Matthieu Puigt}
\ead{matthieu.puigt@univ-littoral.fr}

\cortext[cor1]{Corresponding author}
\affiliation[1]{organization={Univ. Littoral C\^ote d'Opale, LISIC -- UR 4491},
postcode={62219},
city={Longuenesse},
country={France}}

\begin{abstract}
Synthetic spectral image generation is essential for remote sensing simulation and mission design, yet physically based radiative transfer models (RTMs) remain computationally expensive. Existing learning-based emulators reduce this cost, but are mostly deterministic parameter-to-spectrum regressors with limited spatial modeling and uncertainty information. We formulate spectral image emulation as a parameter-conditioned latent-variable problem and propose a variational autoencoder (VAE)-based framework combining nonlinear spectral-image representations, fast inference, and per-pixel uncertainty estimates. The framework is instantiated at spectrum and spatial--spectral levels through two-step VAE pretraining and latent mapping. We evaluate it on PROSAIL-simulated hyperspectral vegetation cubes (211 bands) and real Sentinel-3 OLCI multispectral ocean-colour imagery (21 bands) against classical regression emulators and a deep CNN baseline. Results show that no single architecture is optimal: pixel-to-pixel models perform best on controlled hyperspectral simulations, whereas a fully convolutional VAE is more robust on noisy real observations with missing or contaminated pixels. VAE-based emulators also achieve high throughput for large-scale generation. On Sentinel-3, the spatial--spectral VAE provides predictive intervals closer to empirical errors than pixel-wise neural and classical emulators, although absolute calibration remains incomplete. A look-up-table-based retrieval experiment further shows that reconstruction fidelity alone does not ensure reliable leaf area index and chlorophyll retrieval. Emulators should therefore be evaluated in representative remote-sensing end-use scenarios, not by reconstruction metrics alone.
\end{abstract}


\end{frontmatter}

\input{sections/Introduction}
\input{sections/Section2}

\input{sections/Section3}

\input{sections/Section4}

\input{sections/Section5}

\input{sections/Section6}

\section{Declaration of generative AI and AI-assisted technologies in the manuscript preparation process}

During the preparation of this work, the author(s) used Cursor for proof reading and styling. The author(s) reviewed and edited the output as needed and take full responsibility for the content of the published article.

\section{Data and code availability}

The SVH-BD simulated vegetation hyperspectral benchmark dataset is publicly available on Zenodo~\cite{benazizi2026data} (\url{https://doi.org/10.5281/zenodo.18660571}). Source code, training configurations, and evaluation scripts for reproducing the experiments reported in this article will be released in a public repository concurrent with publication.

\section{Acknowledgements}

This work is partially funded by the ULCO research pole ``Mutations Technologiques et Environnementales``, EUR MAIA (ANR-22-EXES-0009) and Hauts-de-France region.
This work is supported by the graduate school IFSEA that benefits from a France 2030 grant (ANR-21-EXES-0011) operated by the French National Research Agency. Experiments presented in this paper were carried out using the CALCULCO computing platform, supported by DSI/ULCO (Direction des Syst\`emes d'Information de l'Universit\'e du Littoral C\^ote d'Opale).

\appendix
\input{appendix/appendix}

\bibliography{references}

\end{document}

%% file: sections/Introduction.tex
\section{Introduction}

% Context : What is hyperspectral ?
Hyperspectral remote sensing has become an important tool for Earth observation, providing granular insights into environmental processes and land surface characteristics. Hyperspectral imaging acquires dense spectral information across hundreds of contiguous bands for every pixel, yielding a three-dimensional data cube (two spatial dimensions and one spectral dimension) with high spectral resolution. Such data are referred to as hyperspectral images (HSI) and support a wide variety of applications, including vegetation trait retrieval, mineral and soil mapping, water quality and ocean color monitoring \cite{Govender2009,Ghamisi2017}.
% The problem
While collections of high-quality data are continuously produced, their acquisition is often constrained by high costs, atmospheric variability, and limited spatiotemporal coverage. As a result, realistic HSI generation is essential for tasks such as algorithm development and mission design. To this end, simulation-based approaches have been developed to synthetically generate hyperspectral data under controlled conditions. In vegetation remote sensing, the PROSAIL radiative transfer model (RTM)\cite{jacquemoud2009} has proven to be the go to tool to address this challenge. PROSAIL combines the PROSPECT leaf optical properties model\cite{jacquemoud1990prospect} and the SAIL canopy bidirectional reflectance model\cite{verhoef1984light} to simulate top-of-canopy reflectance from plant and soil bio-optical parameters. However, modeling complex non-linear physics through such numerical simulations is computationally expensive, limiting their use in large-scale generation, iterative inversion and operational scenarios.
% Existing solution family : emulation
The recent development of statistical and machine learning techniques has renewed interest in emulation as a framework for synthetic data generation. Emulation involves learning surrogate models that approximate complex simulations with a high degree of accuracy while significantly reducing computational cost. Such approaches are well established in other scientific fields, including astrophysics\cite{guilloteau2020} and climate modeling\citep{yu2023climsim}. 

% Current gap in the literature. Compressed version to match with section II and avoid redundancy.
In hyperspectral imaging, existing emulation approaches include classical regression-based methods based on linear dimensionality reduction, as well as more recent deep-learning-based surrogates for radiative transfer and atmospheric models, which are discussed in detail in Sect.~\ref{sec:relatedwork}. Across these approaches, most existing emulators are formulated as parameter-to-spectrum or parameter-to-band regression models. These formulations typically produce a single deterministic output per parameter configuration, do not learn an explicit distribution of plausible spectral images, and provide limited information about the reliability of the generated data. This is an important limitation for remote-sensing simulation pipelines, where emulated images may be used for downstream tasks rather than visual inspection alone. Despite the widespread use of generative deep learning for hyperspectral image analysis, parameter-conditioned generative modeling for spectral image emulation remains comparatively unexplored. 
%% CITE RECENT ADVANCES IN FM (LIMITATION: DESIGNED FOR DOWNSTREAM TASKS, COMPUTATIONAL COST). ADD RESULTS IF READY
Recent advances in deep generative models for emulation in other scientific fields motivate the investigation of variational autoencoder-based formulations for efficient hyperspectral image emulation.

% Proposed work
Building on our earlier spatial--spectral VAE proof of concept in \cite{benazizi2024}, the present work develops uncertainty-aware latent emulation through a coupled VAE + mapper formulation and studies it across two neural instantiations: a new pixel-to-pixel variant and a fully convolutional spatial--spectral variant. We evaluate the performance of parameter-conditioned spectral image emulation as an uncertainty-aware surrogate for simulation and retrieval workflows. Relative to the preliminary study, we broaden the analysis to include systematic benchmarking against classical and deep regression surrogates, predictive-uncertainty calibration, computational throughput, and downstream biophysical retrieval. The contributions are summarized as follows:

\begin{enumerate}
	\item We propose a latent uncertainty-aware emulation scheme based on coupled VAE + mapper models, instantiated as a spectrum-level pixel-to-pixel architecture and a spatial--spectral fully convolutional architecture, with per-pixel predictive uncertainty estimated through variational latent sampling and MC dropout.
 	\item We benchmark both neural variants against classical regressor + dimensionality reduction (DR) emulators and a deep regression baseline on high-dimensional PROSAIL simulations and real Sentinel-3 OLCI imagery, showing that no single architecture is optimal across data regimes.
	\item We quantify predictive-uncertainty calibration and show that spatial--spectral modeling yields predictive intervals closer to empirical errors on real observations than pixel-wise neural or classical surrogates, despite imperfect absolute calibration.
	\item Through a downstream look-up-table-based biophysical retrieval task, we show that high reconstruction fidelity alone does not guarantee reliable inversion, motivating emulator evaluation in representative end-use scenarios rather than on reconstruction metrics alone.
\end{enumerate}

% Paper Structure
The remainder of the paper is organized as follows. Section~\ref{sec:relatedwork} reviews related work. Section~\ref{sec:methodo} describes the proposed methodology. Section~\ref{sec:expe} reports the experimental setup and results, and Section~\ref{sec:usecase} discusses a practical use case for hyperspectral image emulation.

%% file: sections/Section2.tex
\section{Related work \label{sec:relatedwork}}

\input{sections/Section2/classical_rtm}
\input{sections/Section2/deep_learning_emulators}
\input{sections/Section2/generative_hs}
\input{sections/Section2/generative_beyond_rs}
\input{sections/Section2/positioning}

%% file: sections/Section2/classical_rtm.tex
\subsection{Classical radiative transfer model emulation}
Radiative transfer model (RTM) emulation has been extensively investigated in hyperspectral remote sensing, particularly in the context of vegetation spectroscopy \cite{Verrelst2025}. A representative line of work is presented by Verrelst \emph{et al.} \cite{verrelst2017,verrelst2019,verrelst2021}, who developed several emulators to approximate complex RTMs for tasks such as reflectance or radiance spectra generation, interpolation and extrapolation. These approaches aim to replace computationally expensive numerical simulations with fast surrogate models conditioned on biophysical parameters. In this context, let $\mathbf{x}$ denote a set of scene-related traits (e.g., leaf area index, chlorophyll content in the context of vegetation). Depending on the application, $\mathbf{x}$ can be a vector in $\mathbb{R}^D$, where $D$ is the number of traits, or a set of parameter maps in $\mathbb{R}^{H \times W \times d}$, where H and W are the spatial dimensions. Let $\mathbf{y} \in \mathbb{R}^B$ (or $\mathbb{R}^{H \times W \times B}$ in the spatial-spectral setting) represent the corresponding hyperspectral signal, where $B$ is the number of spectral bands. The forward process can be expressed as:
\begin{equation}
\mathbf{y} = f(\mathbf{x}) + \boldsymbol{\epsilon},
\end{equation}
where $f(\cdot)$ denotes the forward model and $\boldsymbol{\epsilon}$ accounts for measurement and modeling noise. From a probabilistic perspective, this defines a conditional distribution:
\begin{equation}
p(\mathbf{y} \mid \mathbf{x}) = \mathcal{N}(\mathbf{y}, \sigma^2 I),
\end{equation}
assuming additive Gaussian noise with variance $\sigma^2$. The goal of RTM emulation is then to learn a parametric approximation $p_\theta(\mathbf{y} \mid \mathbf{x})$ of this conditional distribution. While efficient and easy to implement, this one step formulation of the emulation problem leads to hidden representations that are hard, if not impossible, to analyze. Concurrently, an intermediate dimension reduction (DR) step can be applied to the target spectra\cite{riveracaicedo2017}, resulting in the following reformulation: 
\begin{equation}
\label{eq:emulation}
\mathbf{y} = D(g(\mathbf{x})) + \boldsymbol{\epsilon},
\end{equation}
where $g(\cdot)$ represents the regression model operating in the latent space, and $D(\cdot)$ reconstructs the hyperspectral data from the latent representation back into the original spectral space.

Such a two step formulation can be found in \cite{verrelst2017}, where an emulator of the SCOPE RTM model was proposed to generate synthetic canopy reflectance and sun-induced fluorescence spectra. First, a PCA is applied to simulated spectra to reduce dimensionality. Then, a regression model is trained to predict the corresponding principal component coefficients from input biophysical parameters. Finally, new spectra are reconstructed from the predicted components. Similar PCA-based emulation strategies were later applied to the emulation of spectral fitting method-based sun-induced fluorescence from HyPlant radiance data \cite{verrelst2019}, and to the interpolation and extrapolation of empirical surface reflectance spectra acquired by CHRIS and HyMap sensors \cite{verrelst2021}. 

These classical emulation approaches have demonstrated that parameter-conditioned emulators can effectively approximate RTMs at a fraction of the computational cost. However, they are primarily designed for spectrum-level emulation and rely on linear dimensionality reduction techniques, such as PCA. While effective for compression, these linear projections remain limited in capturing the intrinsically nonlinear relationships induced by radiative transfer physics \cite{riveracaicedo2017, Laparra2015}. Such non-linearities commonly arise due to effects such as multiple scattering, material mixtures and non-linear spectral mixing processes \cite{altmann2012,dobigeon2014}. Additionally, the methods do not explicitly account for spatial context and are therefore limited in modeling complex spatial-spectral structures inherent to hyperspectral images.

%% file: sections/Section2/deep_learning_emulators.tex
\subsection{Deep learning-based emulators for radiative transfer and atmospheric models}

Beyond classical regression-based emulation, recent studies have explored the use of deep learning models as parameter-conditioned regression surrogates for radiative transfer and atmospheric models. In \cite{duffy2019}, deep Bayesian neural networks were investigated for the emulation of the MAIAC atmospheric correction model. The authors showed that deep learning can capture complex model behavior while significantly reducing execution time compared to traditional lookup-table (LUT) interpolation. Similarly, Ojaghi \emph{et al.} \cite{ojaghi2023} introduced a one-dimensional convolutional neural network (1D-CNN) to emulate the PROSAIL and 6S radiative transfer models for efficient Bidirectional Reflectance distribution function (BRDF) modeling over Sentinel-3 OLCI spectral bands. More recently, Jasso \emph{et al.} \cite{jasso2024} compared different deep learning-based architectures, including multilayer perceptrons (MLPs), numerically embedded MLPs, 1D-CNNs and vision transformers, for the emulation of atmospheric correction from synthetic data generated with the 6S radiative transfer model.

These deep learning-based emulators demonstrate improved accuracy and computational efficiency over classical interpolation and regression techniques, highlighting the ability of neural networks to approximate complex forward models. Nonetheless, they are formulated as direct regression mappings from input parameters to spectral outputs, without explicitly learning a generative latent representation of hyperspectral data. As a result, they are applied at the spectrum or band level and are not designed to generate spatially coherent hyperspectral images.

%% file: sections/Section2/generative_hs.tex
\subsection{Generative models in hyperspectral remote sensing}

While generative models have been largely absent from hyperspectral image emulation, they have been widely investigated in hyperspectral remote sensing for a variety of tasks. In particular, deep generative networks have been employed for data augmentation, classification, spectral unmixing, and spectral variability modeling. This line of research demonstrates the ability of generative models to capture complex hyperspectral data distributions. Generative adversarial networks (GANs) have been proposed for data augmentation in support of hyperspectral image classification \cite{liu2021}, as well as for hyperspectral image generation with controllable spectral variability \cite{palsson2023} or hyperspectral image super-resolution \cite{Shi2022}. Variational autoencoders (VAEs) have also been explored for hyperspectral data analysis, including spectral unmixing \cite{hadi2022}, and endmember library augmentation \cite{borsoi2020}.

Building on these advances, diffusion-based and flow-based generative models have recently emerged as powerful alternatives to GANs and VAEs for high-dimensional data synthesis \cite{ho2020, lipman2023flowmatching}. Diffusion models generate samples through an iterative denoising process that progressively maps noise to the data distribution \cite{ho2020}. Building on the same principle, flow matching enables the training of continuous normalizing flows through direct vector-field regression. This approach avoids the expensive numerical simulations required by earlier differential-equation-based approaches \cite{lipman2023flowmatching}. From a conceptual perspective, this family of methods offers an appealing framework for conditional generation. Their main setback remains their high computational cost, as multiple optimization steps are generally required during inference. This makes them unsuitable for fast emulation.

Although these generative models show strong potential for hyperspectral data generation and analysis, they are formulated for inverse or data-driven tasks, in which generation is conditioned on observed spectra, image content or labels rather than on biophysical and observation parameters. Consequently, the use of generative models for hyperspectral image emulation conditioned on physical parameters remains largely unexplored. 

%% file: sections/Section2/generative_beyond_rs.tex
\subsection{Deep generative models for emulation beyond remote sensing}

Classical and deep learning-based emulators have demonstrated that surrogate models can effectively approximate complex radiative transfer and atmospheric models at a significantly reduced computational cost. In parallel, deep generative networks have shown strong potential as surrogate models for complex physical simulations and have been successfully applied to emulation tasks in other scientific domains, such as astrophysics \cite{guilloteau2020}.

%% file: sections/Section2/positioning.tex
\subsection{Positioning of the present work}

The review above highlights three complementary research directions: classical radiative transfer model emulation based on regression and dimensionality reduction, deep learning-based emulators formulated as direct regression surrogates, and deep generative models applied to hyperspectral data analysis. In addition, generative models have demonstrated their ability to act as efficient surrogates for complex physical simulations in other scientific fields. 

In this work, we position spectral image emulation as a parameter-conditioned latent-variable modeling problem rather than as deterministic regression in a fixed latent space. We adopt a coupled VAE + mapper formulation as the core uncertainty-aware emulator, which augments each emulated output with a per-pixel predictive uncertainty. We instantiate it at both the spectrum and image levels, and evaluate reconstruction, predictive-uncertainty calibration, and downstream biophysical inversion on hyperspectral simulations and real multispectral observations. We additionally test VAE-based dimensionality reduction within classical regressors to assess whether nonlinear latent spaces improve those pipelines. This formulation motivates the methodology presented in the following section.

%% file: sections/Section3.tex
\section{Methodology \label{sec:methodo}}

Building on the identified gap in uncertainty-aware spectral image emulation, we propose a VAE-based latent-space framework designed to generate spectral images from associated biophysical parameters.

\input{sections/Section3/framework}
\input{sections/Section3/variants}

%% file: sections/Section3/framework.tex
\subsection{Variational latent-space emulation framework}

Traditionally, HS image emulation is performed in two stages. First, a dimensionality reduction step is applied to the data, typically using principal component analysis (PCA) on the image spectra. Second, a regression model is trained to predict the projection coefficients of new spectra \cite{verrelst2017}. This framework circumvents the limitations of regression models when applied directly to high-dimensional data such as HSI, reducing the task to predicting a small number of components rather than the entire spectrum or cube. However, relying on linear DR methods disregards the existing non-linearities arising due to effects such as multiple scattering, material mixtures and non-linear spectral mixing processes \cite{altmann2012,dobigeon2014}. This motivates replacing the linear projection with a nonlinear latent representation that can preserve complex spectral interactions while remaining compact enough for efficient emulation.

VAEs provide such a representation by replacing a deterministic projection with a probabilistic latent space. Given a spectral image or spectrum $\mathbf{y}$, an encoder $E_{\psi}$ maps the input to an approximate posterior distribution $q_{\psi}(\mathbf{z}|\mathbf{y})$ over a latent variable $\mathbf{z}$, and a decoder $D_{\theta}$ reconstructs the data through $p_{\theta}(\mathbf{y}|\mathbf{z})$. The encoder and decoder are trained jointly by maximizing the evidence lower bound (ELBO),
\begin{equation}\label{elbo}
    \mathcal{L}({\theta},{\psi},\mathbf{y}^{(i)}) =
    \mathbb{E}_{q_{{\psi}}(\mathbf{z}|\mathbf{y}^{(i)})}
    [\log p_{{\theta}}(\mathbf{y}^{(i)}|\mathbf{z})]
    - D_{KL}\big(q_{{\psi}}(\mathbf{z}|\mathbf{y}^{(i)})||p(\mathbf{z})\big),
\end{equation}
where the first term encourages accurate reconstruction and the second regularizes the latent posterior toward a prior distribution, usually $p(\mathbf{z})=\mathcal{N}(0,I)$ \cite{kingma2013}. The reparameterization trick, $\mathbf{z}=\boldsymbol{\mu}+\boldsymbol{\sigma}\cdot\boldsymbol{\epsilon}$ with $\boldsymbol{\epsilon}\sim\mathcal{N}(0,I)$, enables stochastic latent sampling while preserving differentiability. In the emulation setting, this gives a compact nonlinear embedding of spectral data together with a sampling mechanism that can be exploited for predictive-uncertainty estimation.

\begin{figure}[htb]
    \centering
    \includegraphics[width=.9\linewidth]{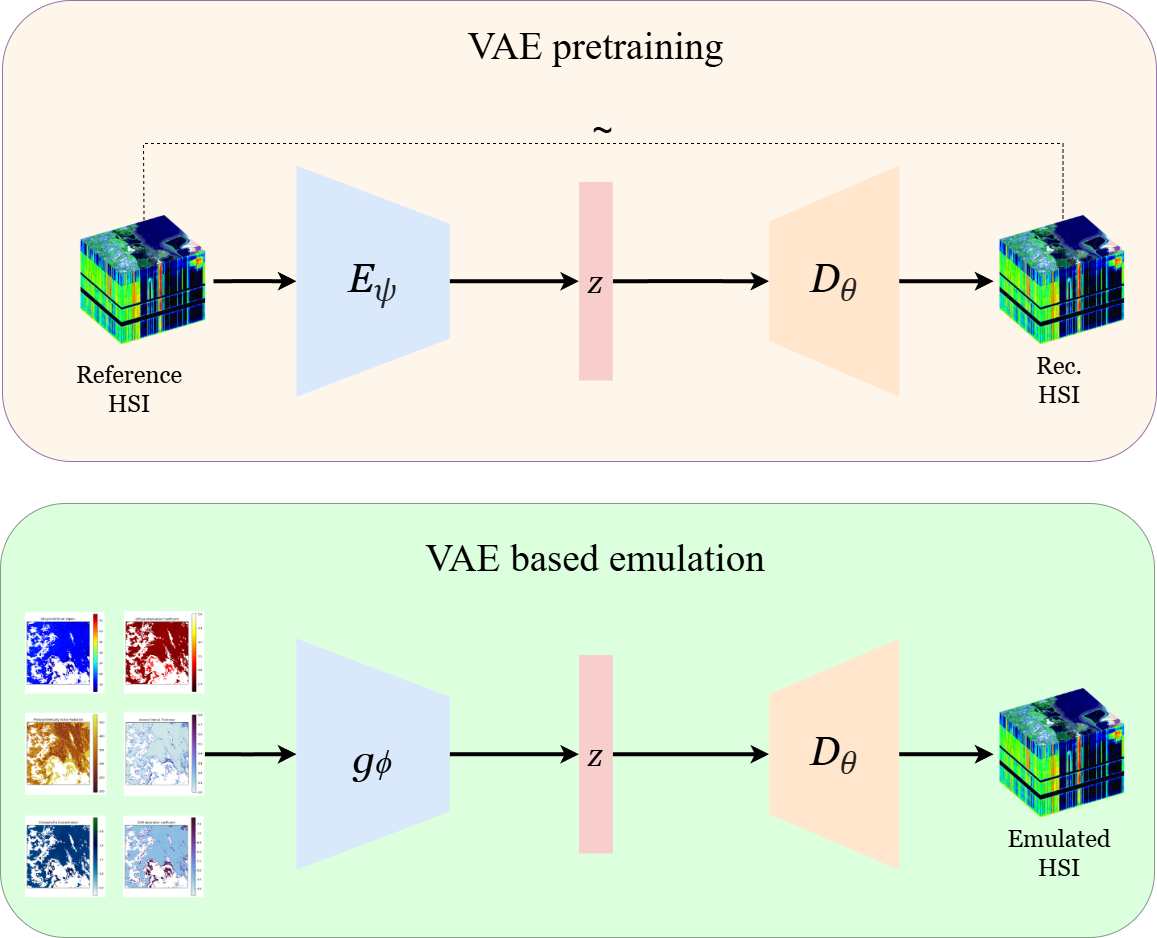}
    \caption{Two-step VAE-based emulation: (top) a VAE is first trained to reproduce hyperspectral images as input. Then, (bottom) a mapper $g_{{\phi}}$  is trained to link the biophysical parameters to the learned latent space.}
    \label{fig:archi}
    % \vspace{-0.25cm}
\end{figure}

We adopt a two-step VAE-based framework inspired by traditional emulation pipelines (Figure~\ref{fig:archi}). We first train a VAE directly on hyperspectral data $\mathbf{y}$ to encode them into a latent representation $\mathbf{z}$, from which the decoder reconstructs the input:
\begin{equation}
\mathbf{z} \sim E_{{\psi}}(\mathbf{y}) = q_{{\psi}}(\mathbf{z}|\mathbf{y}),
\quad \mathbf{\hat{y}} \sim D_{{\theta}}(\mathbf{z}) = p_{{\theta}}(\mathbf{y}|\mathbf{z}),
\end{equation}
where $q_{\psi}$ and $p_{{\theta}}$ denote the approximate posterior and the prior, respectively. Once the VAE is trained and validated for faithful HSI reconstruction, the encoder is discarded. A mapping network $g_\phi$ is then trained to link the biophysical input variables to the latent space:
\begin{equation}
z \sim g_{{\phi}}(\mathbf{x}) = q_{{\phi}}(z|\mathbf{x}),
\quad \mathbf{\hat{y}} \sim D_{{\theta}}(z) = p_{{\theta}}(\mathbf{y}|z),
\end{equation}
where $q_{{\phi}}$ denotes a latent mapping induced by the biophysical parameters, and $p_{{\theta}}$ denotes the decoder distribution. In practice, we write $\mathbf{\hat{y}} = D_{{\theta}}(g_{{\phi}}(\mathbf{x}))$. The parameters ${\phi}$ and ${\theta}$ are optimized to minimize a reconstruction loss between the emulated images $\mathbf{\hat{y}}$ and the reference images $\mathbf{y}$, with an additional Kullback–Leibler divergence term used to regularize the latent distribution:
\begin{equation}
\begin{split}
\min_{{\phi},{\theta}} \mathcal{L}({\theta},{\phi},\mathbf{x})= \min_{{\phi},{\theta}} \Big(\lVert \mathbf{y} - D_{{\theta}}(g_{{\phi}}(\mathbf{x})) \rVert_2^2 + \\ D_{KL}\big(q_{{\phi}}(\mathbf{z}|\mathbf{x})|| p_{{\theta}}(\mathbf{z})\big) \Big).
\end{split}
\label{equation:emulation}
\end{equation}

This two-step approach offers two key advantages. First, the training and validation of each component (VAE for dimensionality reduction and the mapping network) can be carried out sequentially and independently, which facilitates hyperparameter optimization. Second, the model is highly adaptable: any modification of the biophysical input parameters, without altering the properties of the target HSI, only requires retraining the interpolator, leading to significant computational savings.

%% file: sections/Section3/variants.tex
\subsection{Variants}

To evaluate the proposed framework, we implemented two emulator families: a simple pixel-to-pixel baseline, denoted as P2P, and a fully convolutional VAE that incorporates spatial information. We denote the latter as FCVAE.

\paragraph{Pixel-to-pixel emulator (P2P)}  
As a first baseline, we implemented a pixel-to-pixel emulator. The encoder $E_{{\psi}}$, latent mapping network $g_\phi$, and decoder $D_{{\theta}}$ are all implemented as multilayer perceptrons (MLPs). The structures of $E_{{\psi}}$ and $g_\phi$ are identical except for the input layer size, while the decoder $D_{{\theta}}$ mirrors this structure in reverse. Since each pixel (spectrum) is processed independently, the emulator focuses in the spectral domain, at the expense of spatial correlations across the scene. 

\paragraph{Fully convolutional VAE (FCVAE)}  
In the second family (Figure~\ref{fig:archi}), we adopt a fully convolutional design. Both the encoder $E_{{\psi}}$ and latent mapping network $g_\phi$ start with a $1 \times 1$ pointwise convolution to extract spectral features while preserving spatial resolution. This is followed by a set of $7 \times 7$ convolutional layers and  $3 \times 3 \downarrow$ downconvolutional layers. Each layer is succeeded by channel normalization. The exact number of layers are further discussed in section \ref{implementation-details}. The final feature map is projected into $\boldsymbol{\mu}$ and $\log \boldsymbol{\sigma}^2$ through $1 \times 1$ convolutions, representing the mean and log-variance of the approximate posterior $q_{{\psi}}(\mathbf{z}|\mathbf{y})$. The decoder $D_{{\theta}}$ mirrors this structure in reverse. For upsampling, we follow the strategy of \cite{odena2016}: a nearest-neighbor upsampling followed by convolution, which mitigates checkerboard artifacts compared to transpose convolutions. Unlike P2P, the FCVAE directly emulates hyperspectral cubes, thus preserving spatio-spectral coherence.

%% file: sections/Section4.tex
\section{Emulation experiments and results \label{sec:expe}}

This section evaluates the proposed VAE-based hyperspectral emulation framework under different architectural and training configurations. In particular, we study the impact of pixel-wise versus spatial-spectral modeling; experiments use the two-step (pretrained) training strategy, with a direct one-step variant also available. Experiments are conducted on both a simulated vegetation dataset and real Sentinel-3 OLCI imagery to assess model behavior under controlled physical simulation as well as real-world observational conditions. The proposed models are compared against classical regression-based emulators and machine learning baselines commonly used in hyperspectral emulation. We report reconstruction accuracy, spectral fidelity, predictive uncertainty calibration, and runtime performance on CPU/GPU.

\input{sections/Section4/datasets}
\input{sections/Section4/implementation}
\input{sections/Section4/compared_methods}

\subsection{Emulation results}
To assess the performance of the proposed framework, we evaluate reconstruction quality and computational efficiency using complementary metrics. These metrics capture global accuracy, spatial structures preservation, spectral fidelity, and perceptual image quality. 

Reconstruction accuracy is quantified using the Root Mean Squared Error (RMSE) and PSNR, which provide a global measure of emulation accuracy and reconstruction quality. Spatial reconstruction quality is evaluated using the Structural Similarity Index Measure (SSIM) \cite{wang2004}, which compares luminance, contrast, and structural information between reference and emulated images, with higher values indicating better spatial agreement.Spectral fidelity is assessed using the Spectral Angle Mapper (SAM) \cite{yuhas1992}, which measures the angular difference between reference and emulated spectra, and is not sensitive to overall magnitude differences. 

\input{sections/Section4/results_simulated}
\input{sections/Section4/results_real}

\input{sections/Section4/results_interpretation}

\input{sections/Section4/calibration}
\input{sections/Section4/efficiency}

%% file: sections/Section4/datasets.tex
\subsection{Datasets}

We evaluate the proposed framework on two complementary datasets: a simulated vegetation dataset generated with PROSAIL, and real Sentinel-3 OLCI ocean colour imagery. Detailed descriptions of the simulated PROSAIL dataset generation, parameterization, and composition are provided in a separate data publication \cite{benazizi2026svhbd,benazizi2026data}.

\paragraph{Sentinel-3 Ocean Colour}
Sentinel-3 carries multiple instruments, including the Ocean and Land Colour Instrument (OLCI), a pushbroom imaging spectrometer that acquires radiance measurements in 21 bands spanning the visible (VIS) to near-infrared (NIR) range at a spatial resolution of 300~m. The Sentinel-3 OLCI Ocean Colour dataset used in this study consists of radiance images (21 spectral bands, Level-1 products) \cite{s3l1}, together with atmospherically corrected reflectances and associated biophysical parameters related to ocean colour (Level-2 products) \cite{s3l2}. Among the latter, six independent variables were selected as emulator inputs: chlorophyll-a concentration (Chl), total suspended matter concentration (TSM), diffuse attenuation coefficient (Kd), photosynthetically active radiation (PAR), aerosol Ångström exponent (\AA), and integrated water vapour (IWV) \cite{donlon2012}.  

The dataset comprises OLCI acquisitions over several oceanic regions. Each image was cropped into patches of size $128 \times 128$, and only patches dominated by water pixels were retained, with at least 40\% of pixels labeled as water and at most 20\% of pixels belonging to other classes (cloud, land, ice). This filtering resulted in a total of $100\,000$ input–output pairs suitable for training and evaluation.

\paragraph{Simulated PROSAIL dataset : SVH-BD}  
The second dataset consists of $64 \times 64 \times 211$ hyperspectral images with their corresponding bio-optical parameter maps generated by the PROSAIL model \cite{jacquemoud2009}. PROSAIL combines the PROSPECT leaf optical properties model \cite{jacquemoud1990prospect} and the SAIL canopy bidirectional reflectance model \cite{verhoef1984light}, extended with an additional soil bidirectional reflectance distribution function (BRDF) to provide realistic top-of-canopy reflectance $\rho_c$. We use the PROSPECT-D version which accounts for the contribution of brown pigments. At the leaf level, PROSPECT-D is driven by leaf structure, pigment contents (chlorophyll, carotenoids, anthocyanins, and brown pigments), equivalent water thickness, and dry matter content. At the canopy level, SAIL requires parameters describing vegetation structure and viewing geometry, including leaf area index, leaf angle distribution, hotspot effects, and illumination and observation angles. Soil spectra are supplied for each selected region under the assumption of within-region homogeneity, and are drawn from the ICRAF-ISRIC soil spectral library \cite{DVN2021}.   

To construct the simulated dataset, Sentinel-2 multispectral images are first inverted using look-up tables (LUTs) generated with PROSAIL to obtain spatial maps of the bio-optical parameters. These parameter maps provide realistic spatial distributions across the scene and are subsequently used to drive forward PROSAIL simulations, producing pixel-wise canopy reflectance spectra that are rearranged to form hyperspectral cubes. The resulting dataset consists of $10\ 915$ hyperspectral cubes spanning multiple geographic regions, with a ground sampling distance (GSD) of 20 m. These were divided into training, validation, and test subsets.

%% file: sections/Section4/implementation.tex
\subsection{Implementation details}
\label{implementation-details}
For each dataset, the model architectures and training parameters are adapted to the data. Convolutional (FCVAE) models are trained on complete spectral image cubes, whereas pixel-wise (P2P) models are trained on individual spectra extracted from the same scenes. For Sentinel-3, non-water pixels are excluded from pixel-wise training and from the reconstruction evaluation set, as they have no corresponding bio-optical parameters. The Sentinel-3 FCVAE models use $20\ 000$ training patches and $2\ 500$ validation patches, while pixel-wise Sentinel-3 models use approximately $350\ 000$ individual water spectra; a separate set of $2\ 500$ cubes is reserved for testing. For SVH-BD, the $10\ 915$ simulated hyperspectral cubes are split into training, validation, and test subsets using a $70/20/10$ ratio; pixel-wise models use approximately $360\ 000$ spectra sampled from the training scenes.

The FCVAE channel widths are set to 30-60-120-200 for Sentinel-3 and 60-120-200 for SVH-BD. The P2P multilayer perceptrons use hidden layer sizes of 64-128-256 for Sentinel-3 and 256-512 for SVH-BD. All VAE models use cyclic KL annealing to mitigate posterior collapse, with a maximum KL weight of $10^{-3}$, an annealing period of 40 epochs for FCVAE and 10 epochs for P2P. FCVAE and P2P models are trained with initial learning rates of $10^{-3}$ and $10^{-4}$, respectively, followed by exponential decay with $\gamma=0.95$. Dropout with rate $0.1$ is applied to each decoder layer except the output layer. Additional architectural and training details are provided in Appendix~\ref{app:implementation}.

%% file: sections/Section4/compared_methods.tex
\subsection{Compared methods}

We compare the proposed emulation framework against representative regression-based emulators. As classical reference methods, we consider the Kernel Ridge Regression (KRR), Gaussian Process Regression (GPR) and Partial Least Squares (PLS) \cite{verrelst2017,riveracaicedo2017}. These approaches follow the previously introduced two-step emulation framework. First, a dimensionality reduction (DR) technique is applied to the target spectra. Then, a regression model is trained to predict the associated latent components from the biophysical input variables. Each regression method is paired with one of the following DR techniques: Principal Component Analysis (PCA), Canonical Component Analysis (CCA), Minimum Noise Fraction (MNF), Non-Negative Matrix Factorization (NMF), Partial Least Squares (PLS), and a pretrained Variational Autoencoder (VAE DEC) using the same architecture as the P2P-pre pretraining VAE \cite{jolliffe2016pca,wold1966,green1988,hotelling1936} For every combination, several latent-space dimensions were evaluated in order to identify the optimal model configuration. In the presented results, we only show the top two combinations for each regression method. Additional details are provided in Appendix~\ref{app:latent-dimension}.

In addition to these classical approaches, we include a 1D-CNN baseline that directly maps the biophysical input parameters to the full hyperspectral spectrum without explicit dimensionality reduction, using hidden layer sizes of $32$, $64$ and $128$ for both datasets. The 1D-CNN uses the same input variables as the other baselines and is trained using the same optimization settings as the pixel-wise implementations of the proposed framework. This baseline serves to assess the benefit of variational and latent-space modeling compared to a standard feedforward regression approach. All baselines operate at the spectral level and do not explicitly model spatial context. This ensures a fair comparison with the pixel-wise variants of the proposed framework, and highlights the additional benefits of spatial-spectral modeling enabled by the spatial convolutional architectures.

%% file: sections/Section4/results_simulated.tex
\paragraph{Results on the simulated dataset}

\input{tables/metrics_prosail}

The results on the simulated vegetation dataset are reported in Table~\ref{tab:metrics-veg}. Pixel-based neural network approaches, i.e P2P and 1D-CNN, achieve the strongest overall performance in this setting, consistently yielding the lowest reconstruction errors and highest image quality metrics. P2P achieves the best RMSE and PSNR values, while 1D-CNN obtains the highest SSIM and lowest spectral angle, indicating superior structural and spectral fidelity. These results suggest that pixel-based neural architectures are better suited to this reconstruction task than approaches relying on spatial latent representations. On the other hand, FCVAE did not benefit from the added spatial dimension, achieving comparable performance to baseline models.

The results further show that VAE-based latent representations provide a more effective dimensionality reduction framework for classical regression methods than conventional dimensionality reduction techniques. Hybrid approaches based on VAE latent spaces, such as GP + VAE DEC and KRR + VAE DEC, consistently outperform their counterparts relying on classical dimensionality reduction methods in terms of SSIM and SA. This indicates that the learned latent space produced by the VAE preserves structural information more effectively, leading to more performant emulators. Interestingly, among the purely classical approaches, GP + NMF achieves performance comparable with the best methods, even outperforming the FCVAE based emulator. On the other hand, PLS-based methods exhibit substantially weaker reconstruction quality across all metrics.

Qualitative results further corroborate the quantitative findings by highlighting the superior spectral and spatial reconstruction capabilities of neural based approaches. Figure~\ref{fig:scenes-veg} presents a representative emulated scene for the best-performing model of each family (P2P, 1D-CNN, FCVAE, GP + NMF, and GP + VAE DEC), together with selected spectral profiles. P2P and 1D-CNN produce reconstructions that are visually almost indistinguishable from the reference scenes, preserving both spatial coherence and fine-scale vegetation structures without noticeable artifacts. In contrast, spatial latent-space models exhibit slight smoothing effects characteristic of VAE-based generation. The advantage of VAE-based latent representations is also visually apparent when comparing GP + VAE DEC with GP + NMF, as the latter produces substantially noisier reconstructions and reduced spatial consistency.

\input{tables/prosail_rec}

%% file: tables/metrics_prosail.tex
\begin{table}[ht]
    \centering
    \caption{Comparison of evaluated metrics on the simulated dataset. All scores are reported as mean $\pm$ standard deviation over 10 evaluation runs. Best results are shown in bold and second-best results are underlined.}
\hfill
\resizebox{\textwidth}{!}{
\begin{tabular}{lrrrr}
    & \makecell{RMSE $(10^{-2})$ \\ $\downarrow$}
    & SSIM $\uparrow$
    & \makecell{SA $(10^{-2})$ \\ $\downarrow$}
    & \makecell{PSNR \\ (dB)} \\

    \textbf{P2P}
    & $\mathbf{0.60 \pm 0.00}$
    & $\uline{0.9711 \pm 0.0005}$
    & $\uline{2.11 \pm 0.01}$
    & $\mathbf{38.39 \pm 0.04}$ \\
    
    \textbf{FCVAE}
    & $1.17 \pm 0.01$
    & $0.9396 \pm 0.0008$
    & $3.18 \pm 0.03$
    & $32.92 \pm 0.05$ \\
    
    1D-CNN
    & $1.01 \pm 0.01$
    & $\mathbf{0.9897 \pm 0.0001}$
    & $\mathbf{1.80 \pm 0.01}$
    & $35.52 \pm 0.10$ \\

    GPR + NMF
    & $1.04 \pm 0.02$
    & $0.9221 \pm 0.0019$
    & $3.07 \pm 0.17$
    & $34.05 \pm 0.10$ \\

    KRR + NMF
    & $1.43 \pm 0.02$
    & $0.8835 \pm 0.0022$
    & $3.75 \pm 0.19$
    & $31.20 \pm 0.10$ \\

    PLS + NMF
    & $4.38 \pm 0.03$
    & $0.6287 \pm 0.0029$
    & $14.28 \pm 0.49$
    & $20.97 \pm 0.07$ \\

    GP + VAE DEC
    & $1.37 \pm 0.02$
    & $0.9368 \pm 0.0021$
    & $2.82 \pm 0.15$
    & $32.16 \pm 0.10$ \\

    KRR + VAE DEC
    & $1.51 \pm 0.03$
    & $0.9233 \pm 0.0018$
    & $3.29 \pm 0.16$
    & $31.22 \pm 0.10$ \\

    PLS + VAE DEC
    & $4.04 \pm 0.05$
    & $0.6649 \pm 0.0033$
    & $9.00 \pm 0.25$
    & $21.99 \pm 0.10$ \\

\end{tabular}
}
\label{tab:metrics-veg}
\end{table}

%% file: tables/prosail_rec.tex
\begin{figure}[!h]
    \centering
    \newlength{\vegpanelh}
    \setlength{\vegpanelh}{2.5cm}
    % emulated RGB: 500×500; spectra plots: 700×420 — height-only keeps aspect ratio
    \newcommand{\vegscene}[1]{%
        \includegraphics[height=\vegpanelh,keepaspectratio]{#1}%
    }%
    \newcommand{\vegspec}[1]{%
        \includegraphics[height=\vegpanelh,keepaspectratio]{#1}%
    }%
    \setlength{\tabcolsep}{3pt}%
    \renewcommand{\arraystretch}{1}%
    \resizebox{\linewidth}{!}{%
    \begin{tabular}{@{}c@{\hspace{4pt}}c@{\hspace{0.035\linewidth}}c@{\hspace{4pt}}c@{}}
        \multicolumn{2}{c}{\small Reference} & \multicolumn{2}{c}{\small P2P} \\[-2pt]
        \vegscene{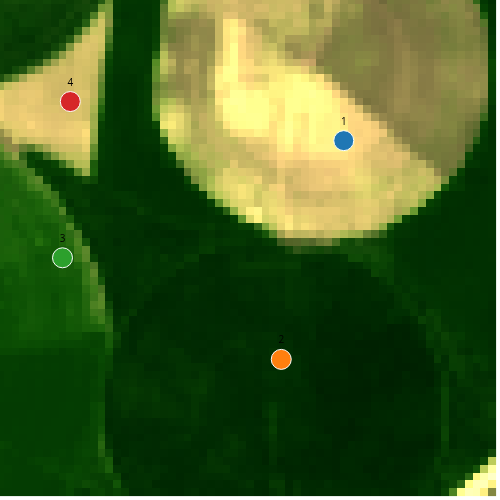} &
        \vegspec{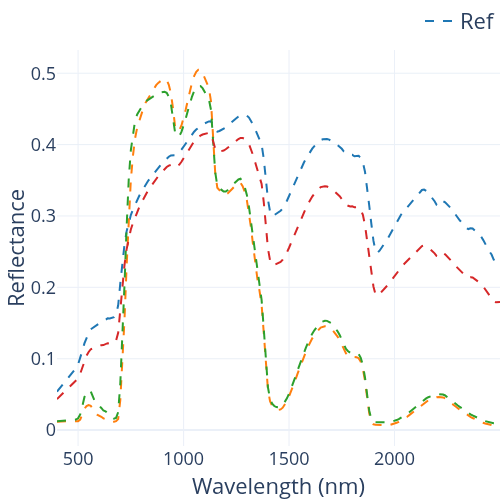} &
        \vegscene{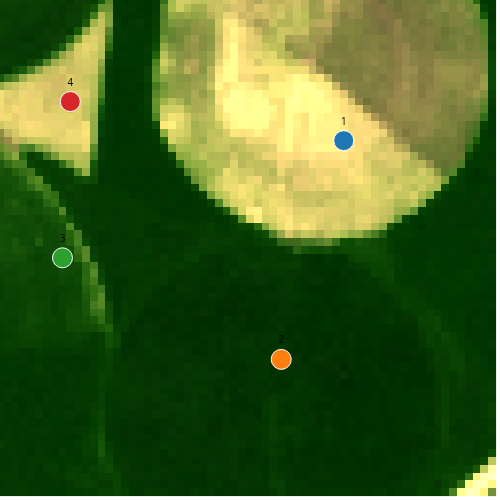} &
        \vegspec{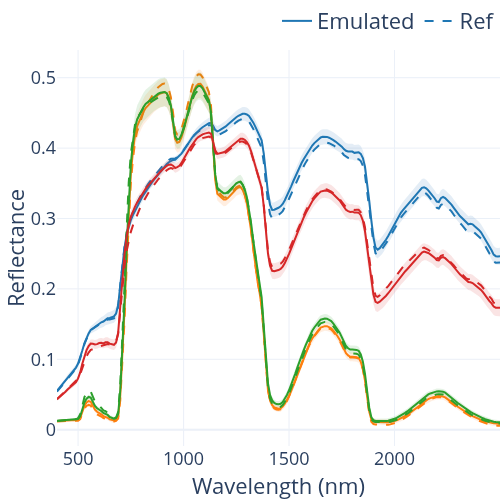}
        \\[8pt]
        \multicolumn{2}{c}{\small 1D-CNN} & \multicolumn{2}{c}{\small FCVAE} \\[-2pt]
        \vegscene{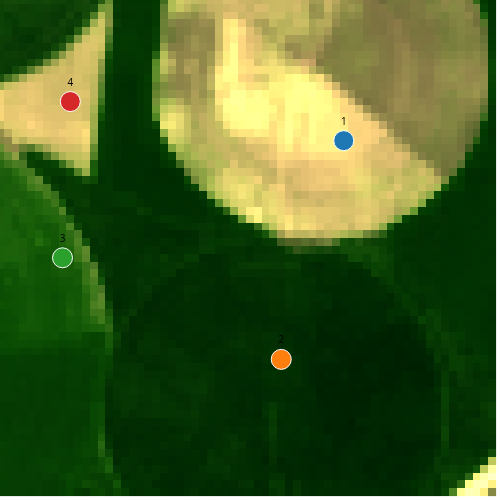} &
        \vegspec{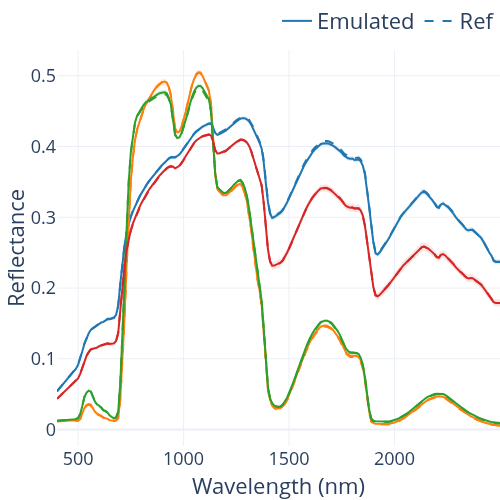} &
        \vegscene{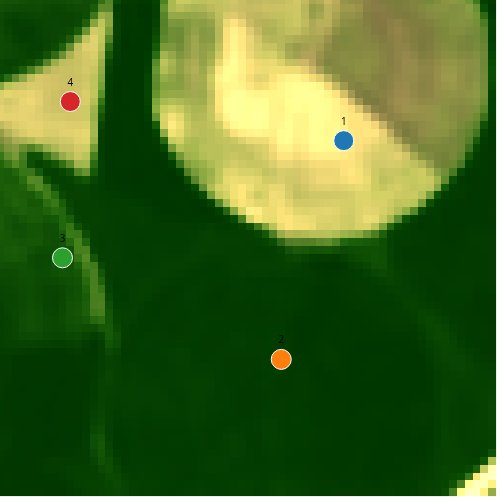} &
        \vegspec{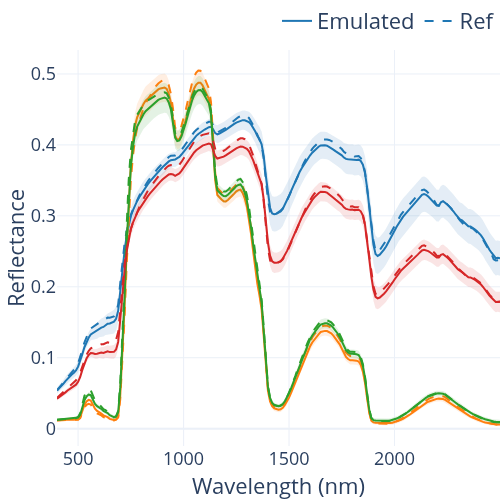}
        \\[8pt]
        \multicolumn{2}{c}{\small GP+NMF} & \multicolumn{2}{c}{\small GP+VAE DEC} \\[-2pt]
        \vegscene{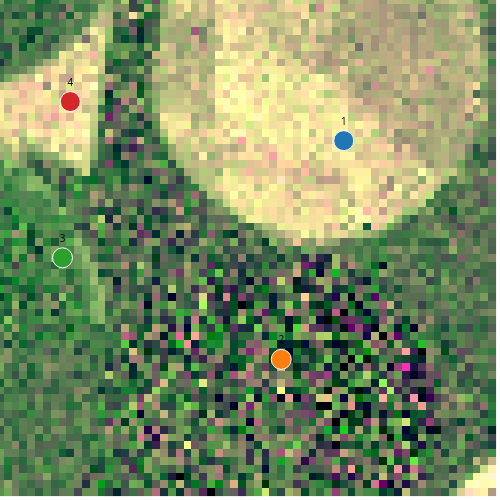} &
        \vegspec{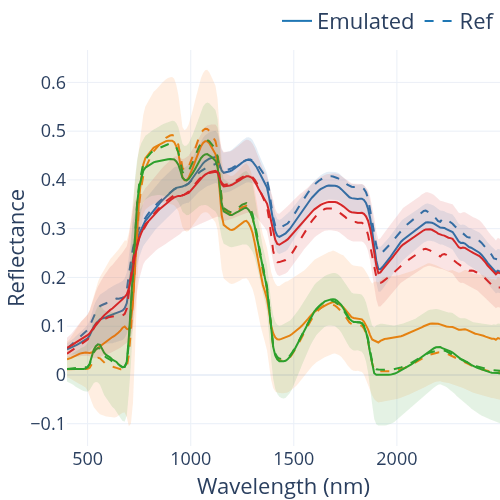} &
        \vegscene{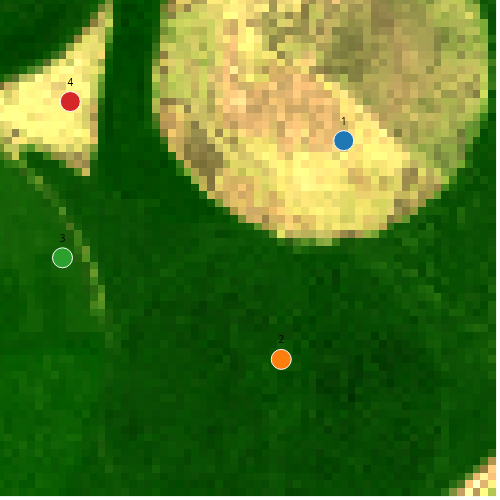} &
        \vegspec{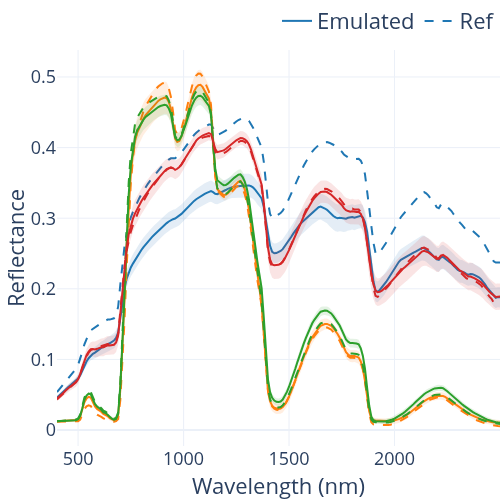}
        \\
    \end{tabular}%
    }

    \caption{Representative PROSAIL-simulated scene for P2P, 1D-CNN, FCVAE, GP+NMF, GP+VAE DEC. Each panel shows an RGB composite (left) and selected spectral profiles with predictive uncertainties (right).}
    \label{fig:scenes-veg}

\end{figure}

%% file: sections/Section4/results_real.tex
\paragraph{Results on the real-world dataset}

\input{tables/metrics_s3}

Results on the Sentinel-3 OLCI dataset are reported in Table~\ref{tab:metrics-s3}.
In contrast to the simulated vegetation case, 1D-CNN achieves the strongest overall performance on real-word imagery across most evaluation metrics, achieving the lowest reconstruction error and the spectral quality. This further highlights the capacity of 1D-CNN architecture at reconstructing realistic spectra. On the other hand FCVAE score the strongest in term of spatial fidelity, all while staying competitive on other metrics. P2P, by contrast, exhibit substantially lower structural similarity and weaker spectral fidelity, although they remain competitive in terms of RMSE and PSNR.

In contrast to the observations on the simulated vegetation dataset, VAE-based latent representations do not systematically improve the performance of classical regression methods on the Sentinel-3 dataset. While KRR + VAE DEC achieves better SSIM than PCA + KRR, the VAE-based variants generally do not consistently outperform their PCA-based counterparts across the remaining metrics. Similarly, GP + VAE DEC remains below GP + PCA in terms of RMSE, SA, and PSNR, despite achieving comparable structural reconstruction quality. These results suggest that, in this context of low-dimensional spectroscopy, the VAE-based latent space does not necessarily provide a more effective representation than conventional linear dimensionality reduction techniques for downstream regression tasks. Among the classical approaches, KRR + PCA achieves the best overall compromise between reconstruction accuracy and spectral consistency, whereas PLS-based methods remain the weakest performers across all evaluation metrics.

Qualitative results provide additional support for these results. As shown in Fig.~\ref{fig:scenes-s3}, pixel-to-pixel emulations on Sentinel-3 imagery exhibit noticeable noise and local inconsistencies, whereas FCVAE produces smoother and more spatially coherent reconstructions. In addition, convolutional models are able to exploit spatial context to plausibly inpaint missing regions, such as land or cloud-contaminated pixels, which contributes to their improved perceptual quality. We note that FCVAE was trained on land and cloud pixels with all input parameters set to zero; we kept the model's predictions for this visualization to illustrate out-of-domain inpainting. 

\input{tables/s3_rec}

%% file: tables/metrics_s3.tex
\begin{table}[ht]
    \centering
    \caption{Evaluation of the emulators on the Sentinel-3 dataset. All scores are reported as mean $\pm$ standard deviation over 10 evaluation runs. Best results are shown in bold and second-best results are underlined.}
    \hfill
\resizebox{\textwidth}{!}{
\begin{tabular}{lrrrr}
    & \makecell{RMSE $\downarrow$}
    & SSIM $\uparrow$
    & \makecell{SA $(10^{-2})$ $\downarrow$}
    & \makecell{PSNR (dB)} $\uparrow$\\
    
    \textbf{P2P}
    & $5.33 \pm 0.04$
    & $0.8555 \pm 0.0013$
    & $4.26 \pm 0.02$
    & $27.54 \pm 0.08$ \\

    \textbf{FCVAE}
    & $\uline{4.77 \pm 0.05}$
    & $\mathbf{0.9160 \pm 0.0010}$
    & $4.19 \pm 0.04$
    & $\uline{28.44 \pm 0.09}$ \\

    1D-CNN
    & $\mathbf{4.20 \pm 0.04}$
    & $0.8662 \pm 0.0014$
    & $\mathbf{2.83 \pm 0.02}$
    & $\mathbf{29.72 \pm 0.06}$ \\

    GP + PCA
    & $6.46 \pm 0.06$
    & $0.7859 \pm 0.0022$
    & $4.50 \pm 0.05$
    & $26.13 \pm 0.08$ \\
    
    KRR + PCA
    & $5.84 \pm 0.06$
    & $0.8287 \pm 0.0019$
    & $\uline{4.06 \pm 0.03}$
    & $26.95 \pm 0.08$ \\

    PLS + PCA
    & $8.43 \pm 0.09$
    & $0.8177 \pm 0.0029$
    & $7.18 \pm 0.08$
    & $23.81 \pm 0.08$ \\
    
    GP + VAE DEC
    & $6.95 \pm 0.06$
    & $0.7846 \pm 0.0021$
    & $5.14 \pm 0.05$
    & $25.31 \pm 0.12$ \\
    
    KRR + VAE DEC
    & $6.41 \pm 0.05$
    & $0.8401 \pm 0.0016$
    & $4.72 \pm 0.03$
    & $26.02 \pm 0.10$ \\

    PLS + VAE DEC
    & $8.83 \pm 0.09$
    & $0.7937 \pm 0.0029$
    & $6.50 \pm 0.04$
    & $23.29 \pm 0.10$ \\

\end{tabular}
}
\label{tab:metrics-s3}
\end{table}

%% file: tables/s3_rec.tex
\begin{figure}[!h]
    \centering
    \newlength{\sThreepanelh}
    \setlength{\sThreepanelh}{2.5cm}
    % emulated RGB: 500×500; spectra plots: 700×420 — height-only keeps aspect ratio
    \newcommand{\sThreescene}[1]{%
        \includegraphics[height=\sThreepanelh,keepaspectratio]{#1}%
    }%
    \newcommand{\sThreespec}[1]{%
        \includegraphics[height=\sThreepanelh,keepaspectratio]{#1}%
    }%
    \setlength{\tabcolsep}{3pt}%
    \renewcommand{\arraystretch}{1}%
    \resizebox{\linewidth}{!}{%
    \begin{tabular}{@{}c@{\hspace{4pt}}c@{\hspace{0.035\linewidth}}c@{\hspace{4pt}}c@{}}
        \multicolumn{2}{c}{\small Reference} & \multicolumn{2}{c}{\small P2P} \\[-2pt]
        \sThreescene{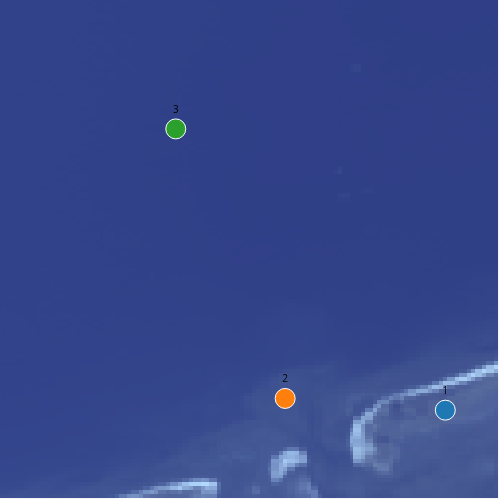} &
        \sThreespec{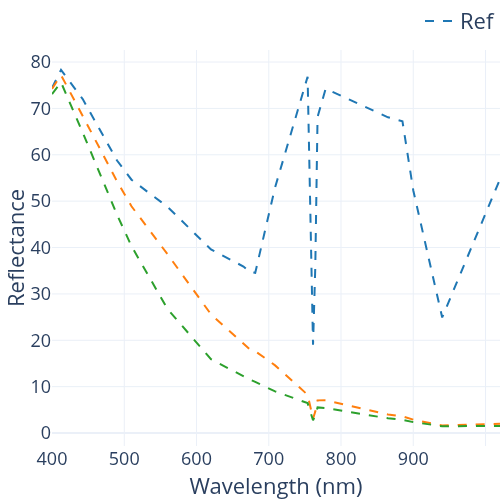} &
        \sThreescene{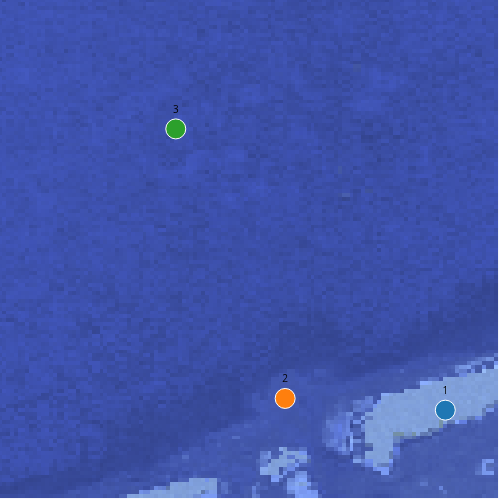} &
        \sThreespec{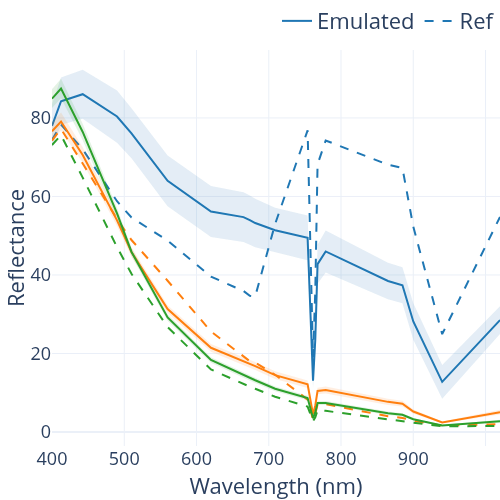}
        \\[8pt]
        \multicolumn{2}{c}{\small 1D-CNN} & \multicolumn{2}{c}{\small FCVAE} \\[-2pt]
        \sThreescene{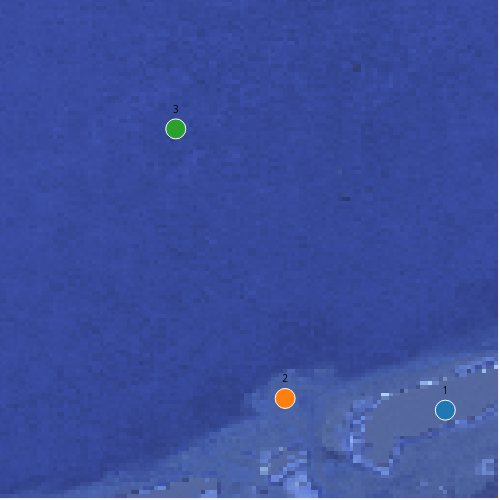} &
        \sThreespec{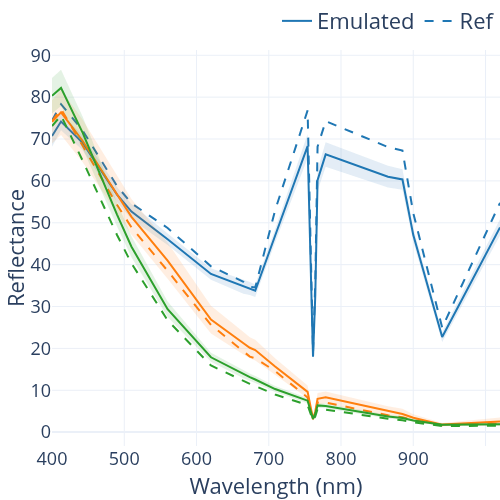} &
        \sThreescene{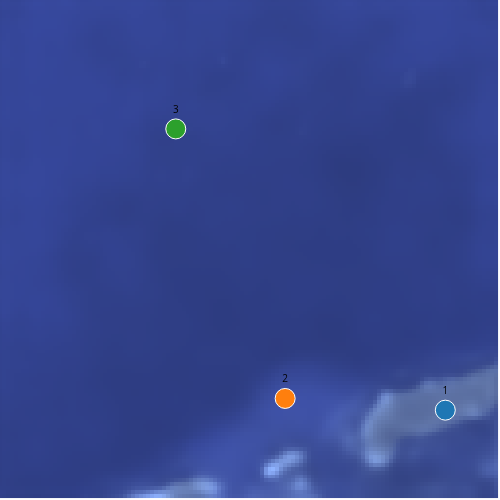} &
        \sThreespec{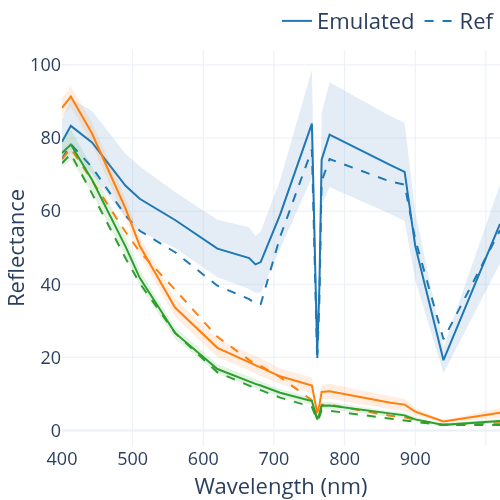}
        \\[8pt]
        \multicolumn{2}{c}{\small KRR+PCA} & \multicolumn{2}{c}{\small KRR+VAE DEC} \\[-2pt]
        \sThreescene{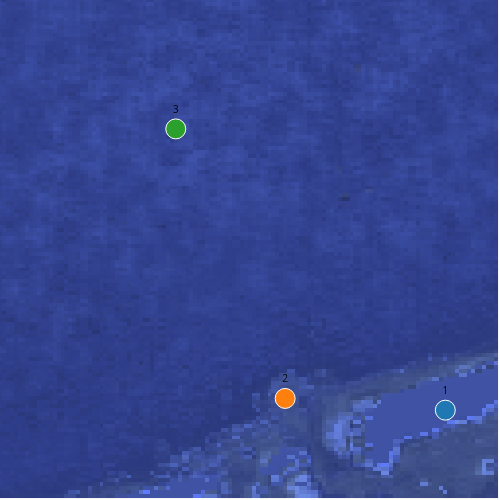} &
        \sThreespec{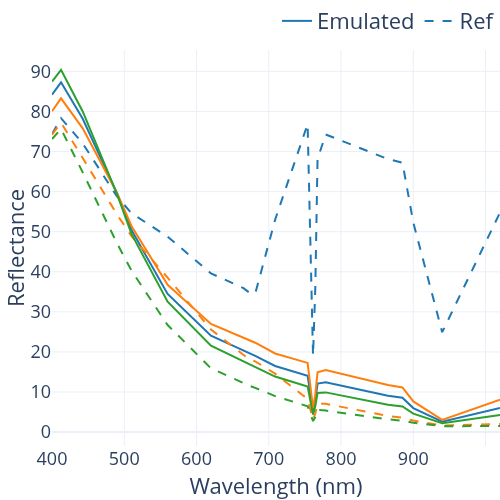} &
        \sThreescene{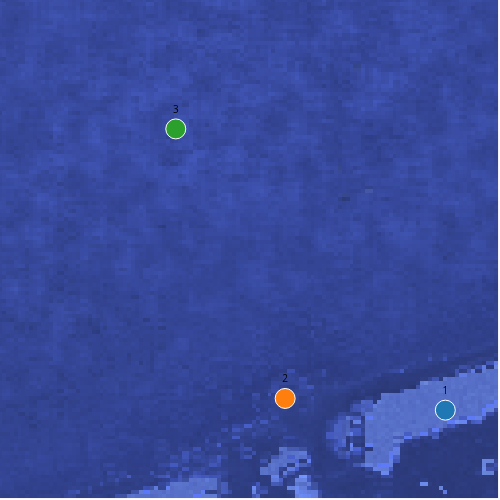} &
        \sThreespec{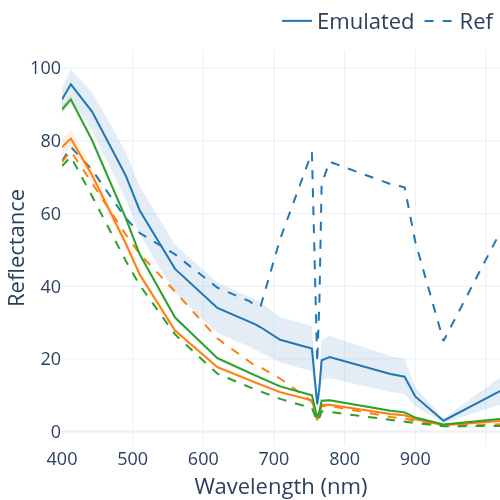}
        \\
    \end{tabular}%
    }

    \caption{Representative Sentinel-3 OLCI patch: reference, P2P, 1D-CNN, FCVAE, KRR+PCA, and KRR+VAE DECODER. Each panel shows an RGB composite (left) and selected spectral profiles with predictive uncertainties (right). FCVAE was trained on land and cloud pixels with all input parameters set to zero; we kept the model's predictions for this visualization to illustrate out-of-domain inpainting.}
    \label{fig:scenes-s3}

\end{figure}

%% file: sections/Section4/results_interpretation.tex
\paragraph{Cross-dataset interpretation}

Taken together, the results reveal a clear dataset-dependent behavior: the models that performs best on high dimensional simulated vegetation data are not necessarily those that perform best on relatively low dimensional ocean satellite imagery. On the simulated vegetation dataset, pixel-to-pixel models achieve the highest reconstruction accuracy because the data are generated spectrum-wise under controlled physical assumptions, with smooth parameter fields, limited noise, and strong spectral consistency. In this context, emulators primarily act as fast surrogates of forward models, and explicit spatial modeling provides limited additional benefit. 

In contrast, real Sentinel-3 imagery is characterized by spatial heterogeneity, sensor noise, and missing or contaminated pixels, which make purely pixel-wise emulation less robust. Under these conditions, accounting for spatial context becomes essential. Convolutional variational models better handle spatial variability by producing more spatially consistent reconstructions and plausibly inpaint missing regions, as observed in the qualitative results, leading to improved robustness on real-world data.

More generally, these observations indicate that there appears to be no universally optimal architecture. Emulator design should therefore be guided by the characteristics of the target data and the intended use case, rather than by performance on simulated benchmarks alone.

%% file: sections/Section4/calibration.tex
\subsection{Uncertainty calibration}
\label{sec:calibration}

Beyond point reconstruction accuracy, we evaluate how informative the predictive uncertainties are for both neural emulators (1D-CNN, P2P, and FCVAE) and representative classical regression baselines. We estimate predictive uncertainty for neural models by Monte Carlo (MC) dropout~\cite{gal2016dropout} and, for VAE-based models, additional latent-space sampling~\cite{kingma2013} (see Appendix~\ref{app:calibration-modes} for latent-only and MC-only ablations). For each model, we draw $S=50$ stochastic forward passes per test band-pixel, estimate a Gaussian predictive distribution from the sample mean and standard deviation, and compute calibration metrics with $K=20$ uncertainty bins. For classical emulators, we use Gaussian-process latent sampling for GP + NMF and GP + VAE DEC on SVH-BD, and  KRR + VAE DEC on Sentinel-3. We assess the calibration of the predictive intervals using several complementary metrics. Empirical coverage measures the proportion of true values contained within the predicted intervals and should match the nominal confidence level (e.g., 90\%)~\cite{gneiting2007calibration}. The root-mean-squared calibration error (RMSCE) quantifies the average deviation between observed and expected coverage across confidence levels, with lower values indicating better calibration. The negative log-likelihood (NLL) evaluates the quality of the predictive Gaussian distribution by jointly assessing calibration and sharpness, where lower values are preferred~\cite{gneiting2007scoring}. Interval sharpness, reported as the mean 90\% interval width, measures the concentration of predictive intervals, with narrower well-calibrated intervals being preferable~\cite{gneiting2007calibration}. The expected calibration error (ECE) compares mean predicted uncertainty and mean absolute error within uncertainty bins. Lower values are preferred for RMSCE, NLL, and ECE, while empirical coverage should closely match the nominal level. Results are summarized in Table~\ref{tab:calibration-summary} and visualized in Appendix~\ref{app:calibration-figures} (Figs.~\ref{fig:calib-reliability-svh}--\ref{fig:calib-coverage-s3}).

\input{tables/calibration_summary}

For classical regression baselines, we use Gaussian-process latent sampling for GP-based pipelines and latent uncertainty from the KRR--DR pipelines. The predictive mean and standard deviation define Gaussian predictive intervals~\cite{gneiting2007calibration}. We assess the calibration of the predictive intervals using several complementary metrics. Empirical coverage measures the proportion of true values contained within the predicted intervals and should match the nominal confidence level (e.g., 90\%)~\cite{gneiting2007calibration}. The root-mean-squared calibration error (RMSCE) quantifies the average deviation between observed and expected coverage across confidence levels, with lower values indicating better calibration. The negative log-likelihood (NLL) evaluates the quality of the predictive Gaussian distribution by jointly assessing calibration and sharpness, where lower values are preferred~\cite{gneiting2007scoring}. Interval sharpness, reported as the mean 90\% interval width, measures the concentration of predictive intervals, with narrower well-calibrated intervals being preferable~\cite{gneiting2007calibration}. The expected calibration error (ECE) compares mean predicted uncertainty and mean absolute error within uncertainty bins. Lower values are preferred for RMSCE, NLL, and ECE, while empirical coverage should closely match the nominal level.

For each model, we draw $S=50$ stochastic forward passes per test band-pixel, estimate a Gaussian predictive distribution from the sample mean and standard deviation, and compute calibration metrics with $K=20$ uncertainty bins. We use combined MC dropout and latent-sampling for VAE-based architectures (see Appendix~\ref{app:calibration-modes} for latent-only and MC-only ablations). For classical emulators, we use Gaussian-process latent sampling for GP + NMF and GP + VAE DEC on SVH-BD, and  KRR + VAE DEC on Sentinel-3. These uncertainty estimates should be interpreted as model-based predictive intervals rather than complete observation-error models. Results are summarized in Table~\ref{tab:calibration-summary} and visualized in Appendix~\ref{app:calibration-figures} (Figs.~\ref{fig:calib-reliability-svh}--\ref{fig:calib-coverage-s3}).

On the simulated SVH-BD dataset, all neural emulators exhibit low calibration errors, as expected under controlled PROSAIL physics and limited sensor noise, but their interval behavior is not identical. The 1D-CNN achieves the lowest RMSCE, NLL, and ECE, but its 90\% predictive intervals are conservative, covering 99.3\% of reference values (nominal 90\%). P2P behaves similarly (99.2\% coverage). FCVAE produces intervals closest to the nominal level (92.6\% coverage), with competitive RMSCE and NLL. The reliability diagrams in Fig.~\ref{fig:calib-reliability-svh} confirm that FCVAE tracks the perfect-calibration diagonal more closely on simulated data, with the smallest coverage gap at the 90\% level. Among classical baselines, GP + VAE DEC achieves low RMSCE and ECE (0.0042 and 0.0035) but under-covers at the 90\% level (62.0\%), whereas GP + NMF is strongly over-conservative (99.8\% coverage) with much wider intervals. In this setting, FCVAE offers the most balanced coverage--sharpness trade-off among the proposed architectures.

The Sentinel-3 OLCI calibration results reveal a markedly different ranking. In contrast to the reconstruction experiment in Table~\ref{tab:metrics-s3}, where non-water pixels were excluded from evaluation, the uncertainty calibration was performed on full test patches without masking land or cloud pixels. This introduces out-of-distribution elements for which no valid bio-optical parameters are available and makes the calibration setting substantially more demanding. Here, 1D-CNN uncertainty estimates are severely overconfident: 90\% intervals cover only 46.6\% of reference band-pixels, with an RMSCE of 5.83 and a NLL of $5.3\times10^{2}$. P2P is similarly miscalibrated (48.6\% coverage, RMSCE 7.32). Both models produce narrow intervals (mean 90\% width 3.5 and 4.7, respectively) that underestimate the true dispersion of errors on real ocean-colour imagery. KRR + VAE DEC is similarly overconfident, reaching only 31.8\% coverage at the 90\% level, with RMSCE of 8.73 and NLL of 162. By contrast, FCVAE yields the predictive intervals closest to empirical errors among the evaluated emulators in this experiment: 69.9\% coverage at the 90\% level, RMSCE 3.77, ECE 1.48, and NLL 3.60---orders of magnitude lower than the pixel-wise neural and classical baselines. This relative improvement comes at the cost of wider intervals (mean 90\% width 9.6 vs.\ 3.5 for 1D-CNN). FCVAE remains under-calibrated in absolute terms (70\% vs.\ 90\% nominal), but it is the only architecture whose predictive intervals are substantially closer to empirical errors on real imagery.

Taken together, these results complement the reconstruction analysis: on simulated vegetation data, pixel-wise regressors dominate accuracy and calibrate easily among neural models. By contrast, on Sentinel-3, the spatial--spectral VAE framework provides per-pixel predictive intervals that are closer to empirical errors than those of both pixel-wise neural emulators and classical regression surrogates, while absolute calibration on real observations remains an open challenge.

%% file: tables/calibration_summary.tex
\begin{table}[ht]
    \centering
    \caption{Predictive uncertainty calibration for neural emulators and classical regression baselines ($S=50$ stochastic forwards or regressor-native samples per input, $K=20$ uncertainty bins). Neural models use the combined MC-dropout and latent-sampling; classical baselines use GP latent sampling (SVH-BD) or KRR latent uncertainty (Sentinel-3). Best results per column and dataset are shown in bold; second-best are underlined. For Sentinel-3, evaluation uses full test patches without water-only masking, unlike Table~\ref{tab:metrics-s3}.}
    \label{tab:calibration-summary}
    \small
    \resizebox{\textwidth}{!}{%
    \begin{tabular}{llrrrrr}
        \hline
        Dataset & Model
        & \makecell{RMSCE\\$\downarrow$}
        & \makecell{NLL\\$\downarrow$}
        & \makecell{Cov.\ 90\%\\$\rightarrow 0.90$}
        & \makecell{Width 90\%\\$\downarrow$}
        & \makecell{ECE\\$\downarrow$} \\
        \hline
        \multirow{5}{*}{SVH-BD}
        & 1D-CNN
        & $\mathbf{0.0013}$ & $\mathbf{-5.36}$ & 0.993 & $\mathbf{0.0063}$ & $\mathbf{0.0012}$ \\
        & P2P
        & 0.0058 & -4.06 & 0.992 & 0.0236 & 0.0048 \\
        & FCVAE
        & $\uline{0.0037}$ & $\uline{-3.65}$ & $\mathbf{0.926}$ & 0.0278 & $\uline{0.0029}$ \\
        & GP + VAE DEC
        & 0.0042 & 0.021 & 0.620 & $\uline{0.0149}$ & 0.0035 \\
        & GP + NMF
        & 0.0407 & -2.38 & 0.998 & 0.138 & 0.0329 \\
        \hline
        \multirow{4}{*}{Sentinel-3}
        & 1D-CNN
        & $\uline{5.83}$ & $5.3\times10^{2}$ & 0.466 & $\mathbf{3.50}$ & $\uline{4.50}$ \\
        & P2P
        & 7.32 & $\uline{83.1}$ & 0.486 & 4.65 & 4.58 \\
        & FCVAE
        & $\mathbf{3.77}$ & $\mathbf{3.60}$ & $\mathbf{0.699}$ & 9.62 & $\mathbf{1.48}$ \\
        & KRR + VAE DEC
        & 8.73 & 162 & 0.318 & $\uline{3.61}$ & 5.99 \\
        \hline
    \end{tabular}%
    }
\end{table}

%% file: sections/Section4/efficiency.tex
\subsection{Computational efficiency}

\begin{figure}[htb]
    \centering
    \includegraphics[width=\linewidth]{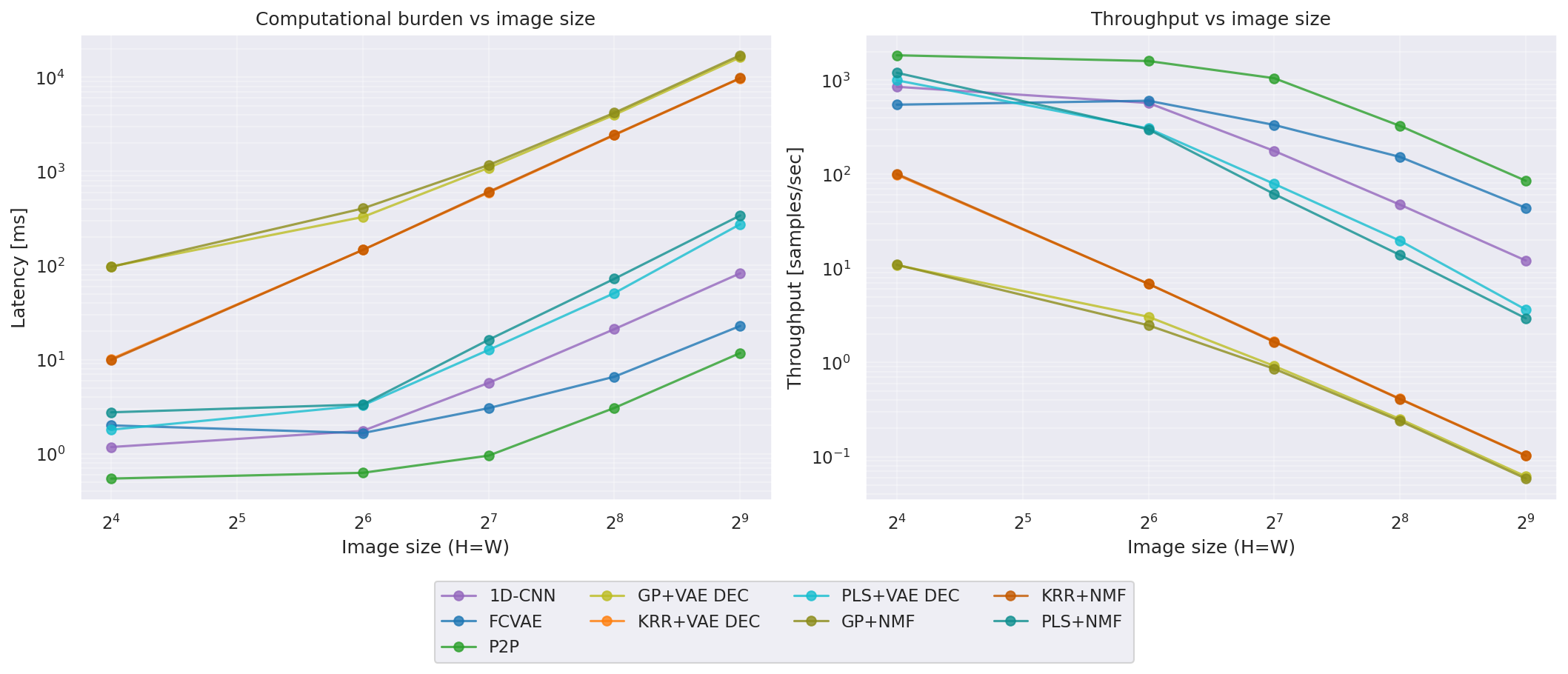}
\caption{Inference throughput and latency measured on an Nvidia A100 40GB GPU for increasing image sizes ($H = W$) with 211 spectral bands.}
    \label{fig:efficiency-cuda}
\end{figure}

\begin{figure}[htb]
    \centering
    \includegraphics[width=\linewidth]{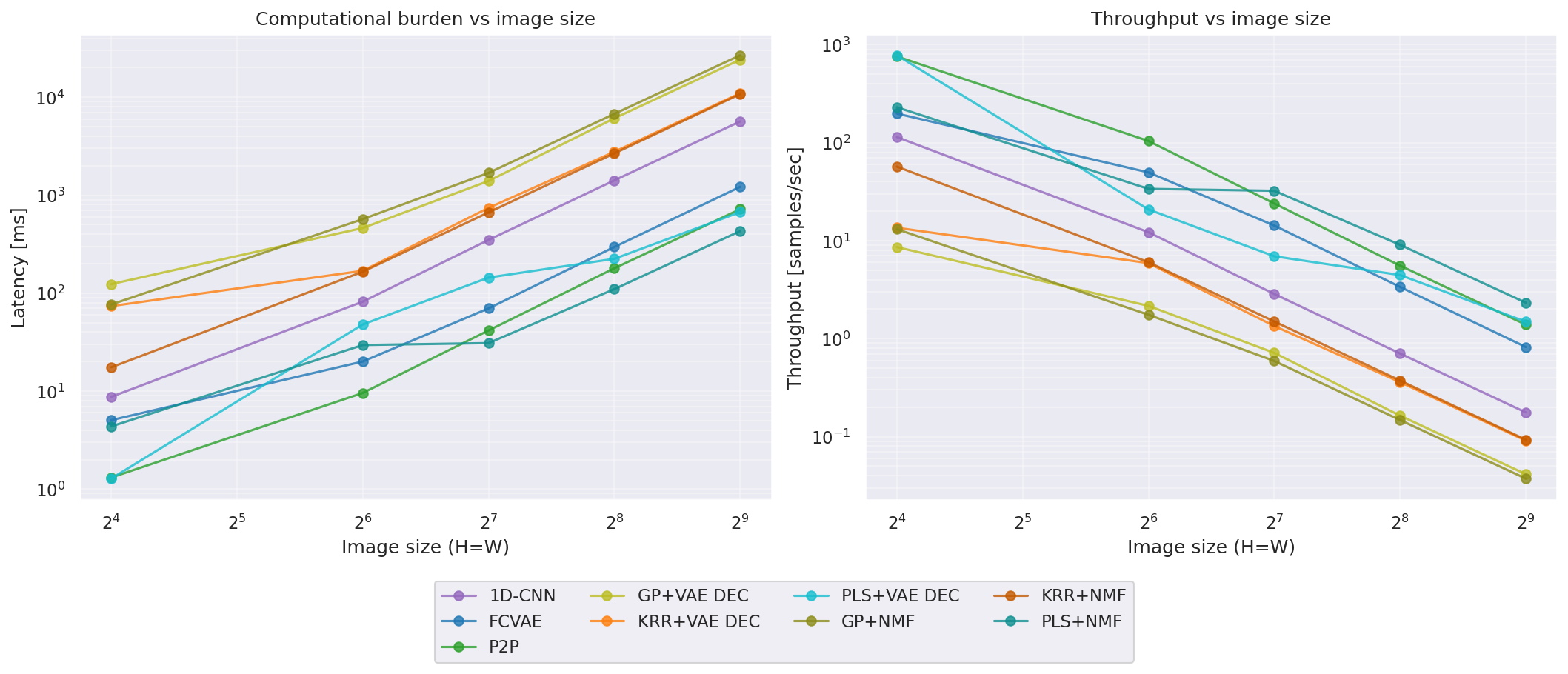}
    \caption{Inference throughput and latency measured on an AMD EPYC 7643 CPU (16 cores) for increasing image sizes ($H = W$) with 211 spectral bands.}
    \label{fig:efficiency-cpu}
\end{figure}

Computational efficiency is presented in Figure~\ref{fig:efficiency-cuda} and Figure~\ref{fig:efficiency-cpu}, which report inference throughput and latency on CPU (AMD EPYC 7643, 16 cores) and GPU (Nvidia A100 40GB) platforms for progressively larger image sizes (with Height ($H$) equal to Width ($W$)). The number of spectral bands is fixed at 211. The results indicate that the lightweight P2P approaches achieve substantially higher speeds than most competing methods, with the exception of the PLS + VAE DEC operating solely on CPU. As anticipated, throughput decreases markedly as image size increases, particularly on CPU, where most models fall below 50 generated samples/s at $H = 2^6$. The decline is less pronounced on GPU hardware, where P2P models still exceed 500 samples/s at $H = 2^8$. FCVAE comes closely behind, at more than 200 samples/s. In contrast, 1D-CNN exhibits even lower throughput and higher latency, less than 100 samples/s at $H = 2^8$. For reference, a LUT-based emulation generates 0.89 images/s at $H = 2^6$, while the full simulator produces one image every 5.58 seconds ($H = 2^6$).

Overall, these results illustrate the expected trade-off between accuracy and efficiency and highlight that computational considerations should be weighed alongside data characteristics when selecting an emulation strategy.

%% file: sections/Section5.tex
\section{Use case for emulated datasets: LAI and Chlorophyll a+b content retrieval
\label{sec:usecase}}

\input{sections/Section5/figures}
\input{sections/Section5/motivation}
\input{sections/Section5/setup}
\input{sections/Section5/retrieval_results}
\input{sections/Section5/implications}

%% file: sections/Section5/figures.tex
\begin{figure}[ht]
    \centering
    \includegraphics[width=0.9\linewidth]{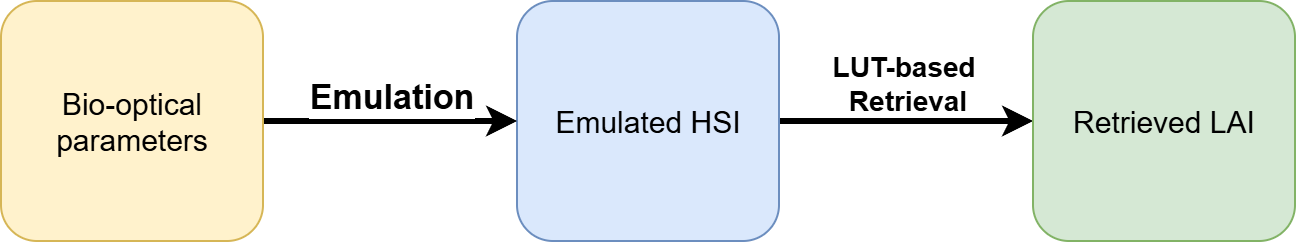}
    \caption{Experimental setup: First, a hyperspectral image is generated using an emulation model. Then we apply LUT-based inversion to retrieve the \textit{Leaf Area Index} and \textit{Chlorophyll a+b content} estimation from the image.}
    \label{fig:lai-retrieval}
\end{figure}

\begin{figure}[p]
\centering
\begingroup
\small
\newcommand{\CabFig}[2]{figures/cab_#1_#2.png}%
\newcommand{\LaiFig}[2]{figures/lai_#1_#2.png}%
\footnotesize
\setlength{\abovecaptionskip}{6pt}%
\subfloat[{\small Ref. $C_{ab}$}]{\includegraphics[width=0.176\linewidth]{\CabFig{GT}{16}}\label{fig:dre:a}}
\hfill
\subfloat[{\small P2P\\RE = 1.78\%}]{\includegraphics[width=0.176\linewidth]{\CabFig{P2P-pre}{16_1.78}}\label{fig:dre:b}}
\hfill
\subfloat[{\small FCVAE\\RE = 2.52\%}]{\includegraphics[width=0.176\linewidth]{\CabFig{FCVAE}{16_2.52}}\label{fig:dre:c}}
\hfill
\subfloat[{\small 1D-CNN\\RE = 2.54\%}]{\includegraphics[width=0.176\linewidth]{\CabFig{1D-CNN}{16_2.54}}\label{fig:dre:d}}
\hfill
\subfloat[{\small GP+VAE DEC\\RE = 19.51\%}]{\includegraphics[width=0.176\linewidth]{\CabFig{DR-GP-pretrained-ld15}{16_19.51}}\label{fig:dre:e}}
\par\vspace{1.2ex}
\subfloat[{\small Ref. LAI}]{\includegraphics[width=0.176\linewidth]{\LaiFig{GT}{16}}\label{fig:dre:f}}
\hfill
\subfloat[{\small P2P\\RE = 1.79\%}]{\includegraphics[width=0.176\linewidth]{\LaiFig{P2P-pre}{16_1.79}}\label{fig:dre:g}}
\hfill
\subfloat[{\small FCVAE\\RE = 2.79\%}]{\includegraphics[width=0.176\linewidth]{\LaiFig{FCVAE}{16_2.79}}\label{fig:dre:h}}
\hfill
\subfloat[{\small 1D-CNN\\RE = 3.98\%}]{\includegraphics[width=0.176\linewidth]{\LaiFig{1D-CNN}{16_3.98}}\label{fig:dre:i}}
\hfill
\subfloat[{\small GP+VAE DEC\\RE = 24.53\%}]{\includegraphics[width=0.176\linewidth]{\LaiFig{DR-GP-pretrained-ld15}{16_24.53}}\label{fig:dre:j}}
\par\vspace{1.2ex}

\subfloat[{\small Ref. $C_{ab}$}]{\includegraphics[width=0.176\linewidth]{\CabFig{GT}{26}}\label{fig:dre:k}}
\hfill
\subfloat[{\small P2P\\RE = 2.06\%}]{\includegraphics[width=0.176\linewidth]{\CabFig{P2P-pre}{26_2.06}}\label{fig:dre:l}}
\hfill
\subfloat[{\small FCVAE\\RE = 2.69\%}]{\includegraphics[width=0.176\linewidth]{\CabFig{FCVAE}{26_2.69}}\label{fig:dre:m}}
\hfill
\subfloat[{\small 1D-CNN\\RE = 3.13\%}]{\includegraphics[width=0.176\linewidth]{\CabFig{1D-CNN}{26_3.13}}\label{fig:dre:n}}
\hfill
\subfloat[{\small GP + VAE DEC\\RE = 23.00\%}]{\includegraphics[width=0.176\linewidth]{\CabFig{DR-GP-pretrained-ld15}{26_23.00}}\label{fig:dre:o}}
\par\vspace{1.2ex}

\subfloat[{\small Ref. LAI}]{\includegraphics[width=0.176\linewidth]{\LaiFig{GT}{26}}\label{fig:dre:p}}
\hfill
\subfloat[{\small P2P\\RE = 1.97\%}]{\includegraphics[width=0.176\linewidth]{\LaiFig{P2P-pre}{26_1.97}}\label{fig:dre:q}}
\hfill
\subfloat[{\small FCVAE\\RE = 3.01\%}]{\includegraphics[width=0.176\linewidth]{\LaiFig{FCVAE}{26_3.01}}\label{fig:dre:r}}
\hfill
\subfloat[{\small 1D-CNN\\RE = 3.30\%}]{\includegraphics[width=0.176\linewidth]{\LaiFig{1D-CNN}{26_3.30}}\label{fig:dre:s}}
\hfill
\subfloat[{\small GP + VAE DEC\\RE = 29.90\%}]{\includegraphics[width=0.176\linewidth]{\LaiFig{DR-GP-pretrained-ld15}{26_29.90}}\label{fig:dre:t}}
\endgroup
\caption{Pixel-wise normalized mean absolute error (relative error, RE) (\%) for LUT-based retrieval on two simulated. For each scene, the first row is $C_{ab}$ and the second is LAI. Columns are, from left to right: reference retrieval from simulated hyperspectral data, P2P, FCVAE, 1D-CNN, and GP + VAE DEC emulations; RE values (\%) are indicated under each emulator panel.}
\label{fig:downstream-retrieval}
\end{figure}

%% file: sections/Section5/motivation.tex
\paragraph{Motivation}
The results presented in Section~\ref{sec:expe} demonstrate that the proposed emulation models are able to reproduce hyperspectral data with high spectral and spatial fidelity. However, reconstruction accuracy alone does not necessarily guarantee that emulated data are suitable for practical remote sensing applications. In particular, small spectral or spatial discrepancies may be propagated when emulated hyperspectral images are used as inputs to nonlinear inversion or retrieval algorithms. To assess the practical reliability of the proposed emulation framework, we therefore evaluate its impact on a representative downstream task, namely the retrieval of biophysical parameters using a look-up table (LUT)-based inversion approach.

%% file: sections/Section5/setup.tex
\paragraph{Experimental setup}

To evaluate the impact of emulated hyperspectral data on downstream biophysical parameter retrieval, we consider a LUT-based inversion for \textit{Chlorophyll a+b} ($C_{ab}$) and for \textit{Leaf Area Index} (LAI). The LUTs are generated from simulated PROSAIL data and are applied to hyperspectral emulated images produced by P2P, FCVAE-pre, 1D-CNN, and GP + VAE DEC emulators. For this use case, we built new look-up tables dedicated to the retrieval experiment, distinct from those used to generate the hyperspectral dataset on which the emulators were trained, so as to avoid data leakage between training data and evaluation. The resulting $C_{ab}$ and LAI maps are compared to reference values obtained from the original hyperspectral simulated data. Figure~\ref{fig:lai-retrieval} illustrates the overall experimental process.

%% file: sections/Section5/retrieval_results.tex
\paragraph{Results on biophysical parameter retrieval}

Figure~\ref{fig:downstream-retrieval} summarizes pixel-wise retrieval errors for two simulated scenes, for both $C_{ab}$ and LAI, and for the same four emulators. On the top, P2P attains the lowest map-scale errors for $C_{ab}$ (1.78\%) and LAI (1.79\%), followed closely by FCVAE (2.52\% and 2.79\%) and the 1D-CNN (2.54\% and 3.98\%), whereas GP + VAE DEC reaches 19.51\% and 24.53\%, respectively. The bottom scene exhibits the same ordering: P2P remains best for $C_{ab}$ and LAI (2.06\% and 1.97\%), with FCVAE and the 1D-CNN again forming a tight cluster (2.69--3.13\% for $C_{ab}$; 3.01--3.30\% for LAI), while GP + VAE DEC yields substantially larger errors (23.00\% and 29.90\%). Across both scenes and variables, the three neural emulators preserve retrieval patterns comparable to the reference maps, with errors concentrated mainly at parcel boundaries, whereas GP + VAE DEC produces stronger spatial artifacts and larger relative errors, which are amplified by the nonlinear LUT inversion. These results indicate that emulation choice strongly conditions downstream biophysical retrieval, even when spectral reconstruction scores are favorable for several models.

%% file: sections/Section5/implications.tex
\paragraph{Implications for practical use of emulated data}
These results highlight that high reconstruction accuracy does not automatically translate into reliable performance in downstream applications, and that the choice of an emulation strategy can have a significant impact on inversion outcomes. While pixel-wise and latent-space learning-based emulators preserve biophysical parameter retrieval accuracy within a narrow margin relative to reference simulations, the GP + VAE DEC pathway can introduce spatially structured errors that are amplified by the nonlinear inversion. This observation underscores the importance of evaluating hyperspectral emulators not only in terms of spectral or spatial fidelity, but also with respect to their behavior in representative end-use scenarios.

%% file: sections/Section6.tex
\input{sections/Section6/discussion}
\input{sections/Section6/conclusion}

%% file: sections/Section6/discussion.tex
\section{Summary and discussion\label{sec:discussion}}

\paragraph{Architecture-dependent behavior}
First, the choice of architecture strongly depends on the application domain. Pixel-to-pixel (P2P) models achieve the highest fidelity on simulated vegetation data, with strong preservation of spectral signatures. Several factors may explain this behavior: (i) the dimensionality reduction is less severe, limiting information loss; (ii) representing independent spectra in a compact latent space is simpler than encoding full hyperspectral cubes; and (iii) learning local spectral dependencies is inherently easier in the case of isolated spectra. These properties collectively account for the superior performance of P2P approaches on simulated vegetation data. By contrast, Sentinel-3 data exhibit greater homogeneity across scenes. In this setting, pixel-to-pixel emulation tends to produce noisier reconstructions. Fully convolutional VAEs perform better by explicitly modeling spatio-spectral correlations, thereby yielding smoother and more coherent outputs. They can also inpaint missing regions, such as land and cloud pixels.

\paragraph{Predictive uncertainty}
The calibration analysis shows that reconstruction accuracy alone does not fully characterize emulator performance. On simulated data, neural models produce low calibration errors, although some intervals are conservative. Clearer differences emerge on real Sentinel-3 imagery. Pixel-wise neural architectures and KRR surrogates become overconfident under distribution shift, whereas the fully convolutional VAE provides the most reliable---but still under-calibrated---uncertainty estimates among the evaluated emulators by leveraging spatial context and probabilistic latent representations (Appendix~\ref{app:calibration-figures}, Figs.~\ref{fig:calib-reliability-s3} and~\ref{fig:calib-coverage-s3}). On Sentinel-3, FCVAE reaches only 70\% empirical coverage at the 90\% nominal level. This gap is mainly attributed to the presence of out-of-distribution pixels and unmodeled error sources, such as sensor and atmospheric effects, which increase the true error variability beyond the uncertainty captured by stochastic latent sampling and dropout. Further calibration strategies could improve the agreement between predicted and empirical uncertainties.

\paragraph{Computational efficiency}
The computational analysis demonstrates that neural emulators benefit substantially from GPU acceleration, making them well suited for large-scale hyperspectral emulation. Among them, our proposed approaches achieve the highest throughput at scale, highlighting their ability to efficiently process large image volumes while maintaining strong reconstruction performance. These results emphasize that computational efficiency is a key consideration for operational remote sensing applications, alongside reconstruction accuracy and uncertainty estimation.

\paragraph{Practical implications and limitations}
Third, our evaluation against classical regression-based emulators highlights differences in both accuracy and robustness. While methods such as KRR and GPR remain competitive in CPU-limited scenarios, they consistently underperform in terms of reconstruction accuracy. However, when annotated data is scare, classical regression methods remain viable options. By contrast, the proposed VAE-based framework offers a favorable trade-off between accuracy and computational efficiency, particularly when executed on GPU architectures, at the cost of requiring big volumes of data. In addition, the practical usability of the emulated data was assessed through a downstream application, namely LUT-based retrieval of \textit{Chlorophyll a+b} ($C_{ab}$). Pixel-to-pixel emulators achieved the best retrieval performance, closely followed by the 1D-CNN. These results highlight the potential of emulated hyperspectral data for downstream applications, while also underscoring the need for careful evaluation of model-induced biases when emulation outputs are used in nonlinear inversion workflows.

%% file: sections/Section6/conclusion.tex
\section{Conclusion and perspectives \label{sec:conclu}}

% Contribution
In this work, we studied fast and uncertainty-aware spectral image emulation using VAE-based pixel-to-pixel and fully convolutional variants. Experiments on simulated PROSAIL hyperspectral data and real Sentinel-3 OLCI multispectral imagery show that emulator design should be guided by target data characteristics and intended use case. Pixel-to-pixel models achieve the highest reconstruction fidelity in controlled simulations, whereas fully convolutional VAEs better handle real-world spatial heterogeneity, sensor noise, and missing pixels while providing the most reliable predictive intervals among the evaluated models on satellite observations, despite remaining under-calibrated in absolute terms. A downstream LUT-based retrieval experiment further demonstrates that reconstruction quality alone does not ensure reliable biophysical inversion. These results support uncertainty-aware, architecture-informed emulator selection for remote sensing simulation pipelines rather than a single universally optimal surrogate.

% Perspectives
Moving forward, future work will focus on latent-space analysis and construction for better interpretability and data representation within the proposed emulation framework. In a second stage, we will investigate the potential of our framework for solving the inverse problem, \textit{i.e} parameter retrieval.

%% file: appendix/appendix.tex
\input{appendix/implementation_details}
\input{appendix/calibration_modes}

\input{appendix/calibration_figures}
\input{appendix/latent_dimension}

%% file: appendix/implementation_details.tex
\section{Implementation details}
\label{app:implementation}

\paragraph{Dataset splits}
For the Sentinel-3 dataset, hyperspectral patches are divided into training and validation subsets, with $20\ 000$ samples used for training and $2\ 500$ for validation in the FCVAE implementations. Pixel-wise (P2P) models are trained and validated on approximately $350\ 000$ individual spectra extracted from the same dataset, while a separate set of $2\ 500$ hyperspectral cubes is reserved for testing. Non-water pixels were removed for the pixel-wise models as they do not have corresponding bio-optical parameters, and were also excluded from the evaluation set. For the simulated PROSAIL dataset, the $10\ 915$ hyperspectral cubes are split into training ($70\%$), validation ($20\%$), and test ($10\%$) subsets. Convolutional models are trained on the full training split, while pixel-wise models are trained at the spectral level using approximately $360\ 000$ individual spectra.

\paragraph{Network dimensions}
For the FCVAE variants, convolutional layer widths are set to 30-60-120-200 for the Sentinel-3 dataset and to 60-120-200 for the simulated dataset. For the pixel-wise implementations, multilayer perceptrons with hidden layer sizes of 64-128-256 are used for the Sentinel-3 dataset and 256-512 for the simulated one.

\paragraph{Training hyperparameters}
To reduce KL divergence vanishing during VAE training, a cyclic annealing strategy is applied with a maximum KL weight of $10^{-3}$. The annealing period is fixed at 40 epochs for FCVAE models and 10 epochs for P2P models. FCVAE and P2P models are trained with initial learning rates of $10^{-3}$ and $10^{-4}$, respectively, which decrease exponentially with a decay rate of $\gamma = 0.95$. Finally, a dropout rate of $0.1$ is applied to each decoder layer except the output layer.

%% file: appendix/calibration_modes.tex
\section{Uncertainty calibration: stochasticity modes}
\label{app:calibration-modes}

Table~\ref{tab:calibration-svh-modes} and Table~\ref{tab:calibration-s3-modes} report the full calibration ablation over three stochasticity sources for each neural emulator: latent sampling only (\texttt{latent}), Monte Carlo (MC) dropout only (\texttt{mc}), and the combined (\texttt{mc+latent}). All runs use $S=50$ stochastic forward passes per evaluated band-pixel and $K=20$ uncertainty bins for binned calibration metrics. The 1D-CNN baseline does not admit latent sampling.

\input{tables/calibration_modes_appendix}

%% file: tables/calibration_modes_appendix.tex
% Appendix: full uncertainty-mode ablation (latent / mc / mc+latent)
% Include via \input{tables/calibration_modes_appendix} in appendix

\begin{table}[htb]
    \centering
    \caption{Uncertainty calibration on the simulated SVH-BD dataset for all stochasticity modes ($S=50$, $K=20$ bins).}
    \label{tab:calibration-svh-modes}
    \scriptsize
    \resizebox{\textwidth}{!}{%
    \begin{tabular}{llrrrr}
        \hline
        Mode & Model
        & RMSCE & NLL & Cov.\ 90\% & ECE \\
        \hline
        \multirow{2}{*}{Latent}
        & P2P & 0.0006 & -3.68 & 0.833 & 0.0003 \\
        & FCVAE & 0.0051 & 1.28 & 0.447 & 0.0046 \\
        \hline
        \multirow{3}{*}{MC}
        & 1D-CNN & 0.0013 & -5.36 & 0.993 & 0.0012 \\
        & P2P & 0.0050 & -4.21 & 0.973 & 0.0037 \\
        & FCVAE & 0.0032 & -3.64 & 0.897 & 0.0022 \\
        \hline
        \multirow{2}{*}{MC+latent}
        & P2P & 0.0058 & -4.06 & 0.992 & 0.0048 \\
        & FCVAE & 0.0037 & -3.65 & 0.926 & 0.0029 \\
        \hline
    \end{tabular}%
    }
\end{table}

\begin{table}[htb]
    \centering
    \caption{Uncertainty calibration on the Sentinel-3 OLCI dataset for all stochasticity modes ($S=50$, $K=20$ bins). Evaluation uses full test patches without water-only masking.}
    \label{tab:calibration-s3-modes}
    \scriptsize
    \resizebox{\textwidth}{!}{%
    \begin{tabular}{llrrrr}
        \hline
        Mode & Model
        & RMSCE & NLL & Cov.\ 90\% & ECE \\
        \hline
        \multirow{2}{*}{Latent}
        & P2P & 6.90 & $1.3\times10^{3}$ & 0.226 & 6.10 \\
        & FCVAE & 4.74 & 80.5 & 0.387 & 3.17 \\
        \hline
        \multirow{3}{*}{MC}
        & 1D-CNN & 5.83 & $5.3\times10^{2}$ & 0.466 & 4.50 \\
        & P2P & 9.24 & $1.2\times10^{2}$ & 0.288 & 5.15 \\
        & FCVAE & 3.47 & 4.08 & 0.671 & 1.56 \\
        \hline
        \multirow{2}{*}{MC+latent}
        & P2P & 7.32 & 83.1 & 0.486 & 4.58 \\
        & FCVAE & 3.77 & 3.60 & 0.699 & 1.48 \\
        \hline
    \end{tabular}%
    }
\end{table}

%% file: appendix/calibration_figures.tex
\section{Calibration reliability diagrams}
\label{app:calibration-figures}

\begin{figure}[!ht]
    \centering
    \subfloat[1D-CNN]{\includegraphics[width=0.24\linewidth]{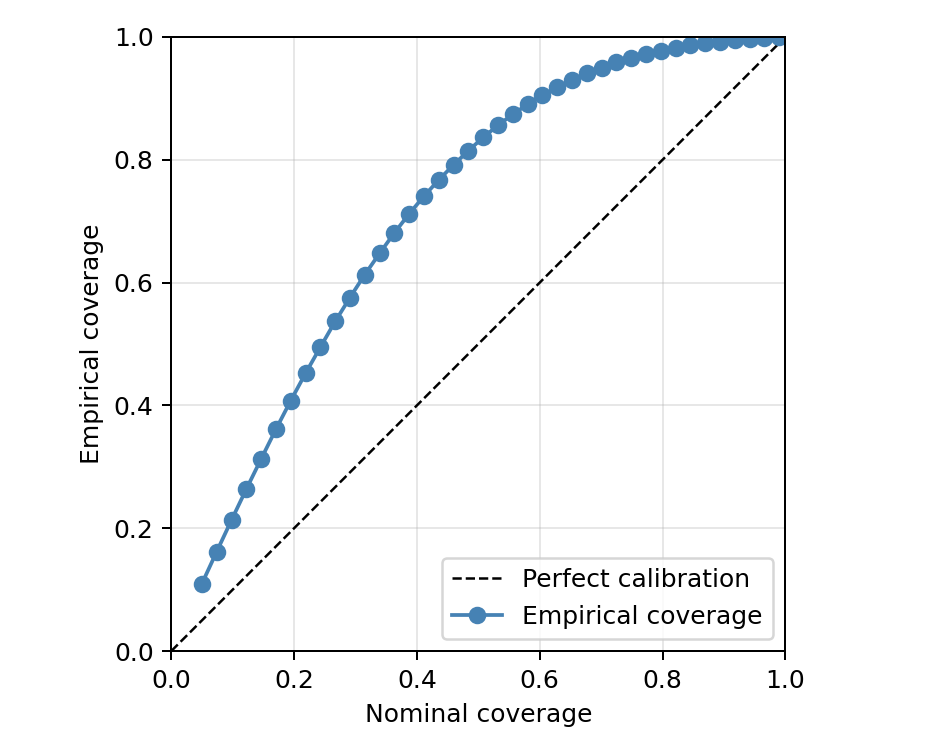}\label{fig:calib-reliability-svh-cnn}}
    \hfill
    \subfloat[P2P]{\includegraphics[width=0.24\linewidth]{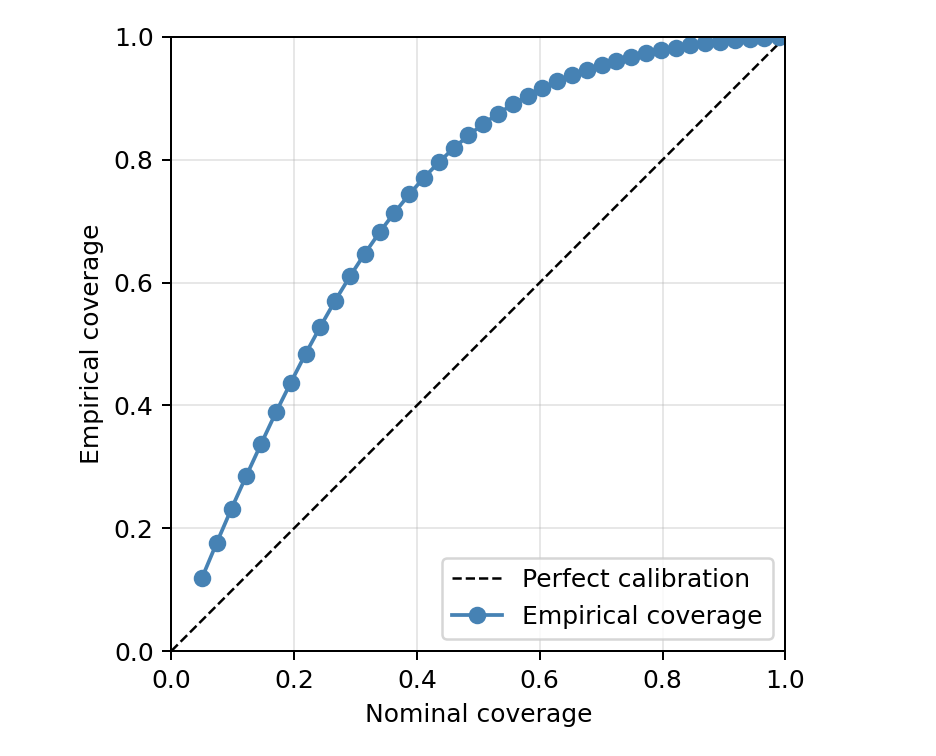}\label{fig:calib-reliability-svh-p2p}}
    \hfill
    \subfloat[FCVAE]{\includegraphics[width=0.24\linewidth]{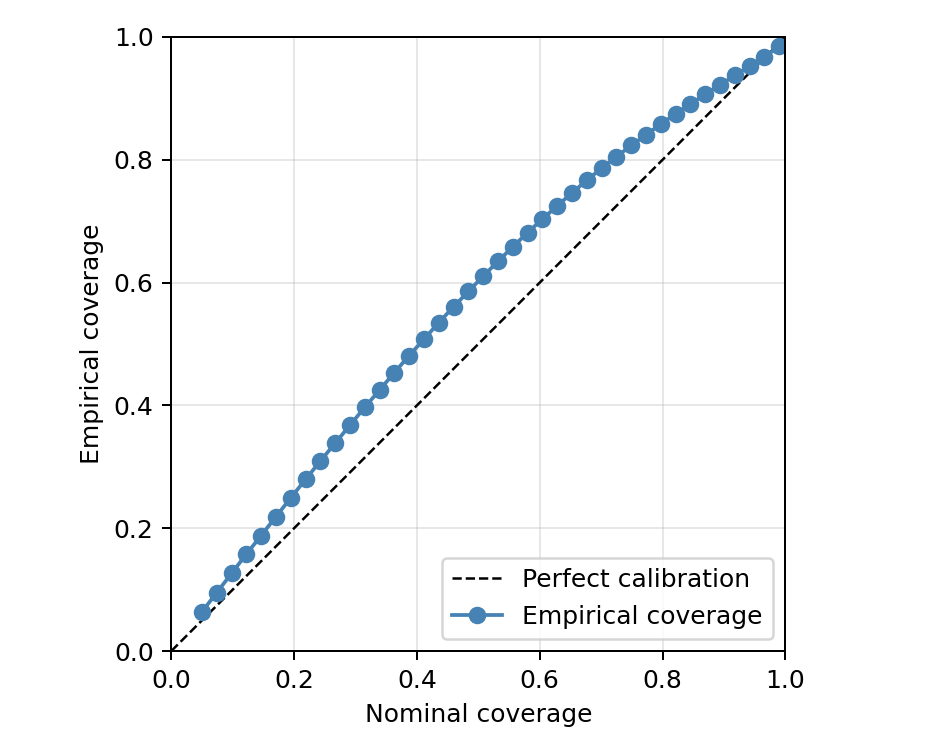}\label{fig:calib-reliability-svh-fcvae}}
    \hfill
    \subfloat[GP + VAE DEC]{\includegraphics[width=0.24\linewidth]{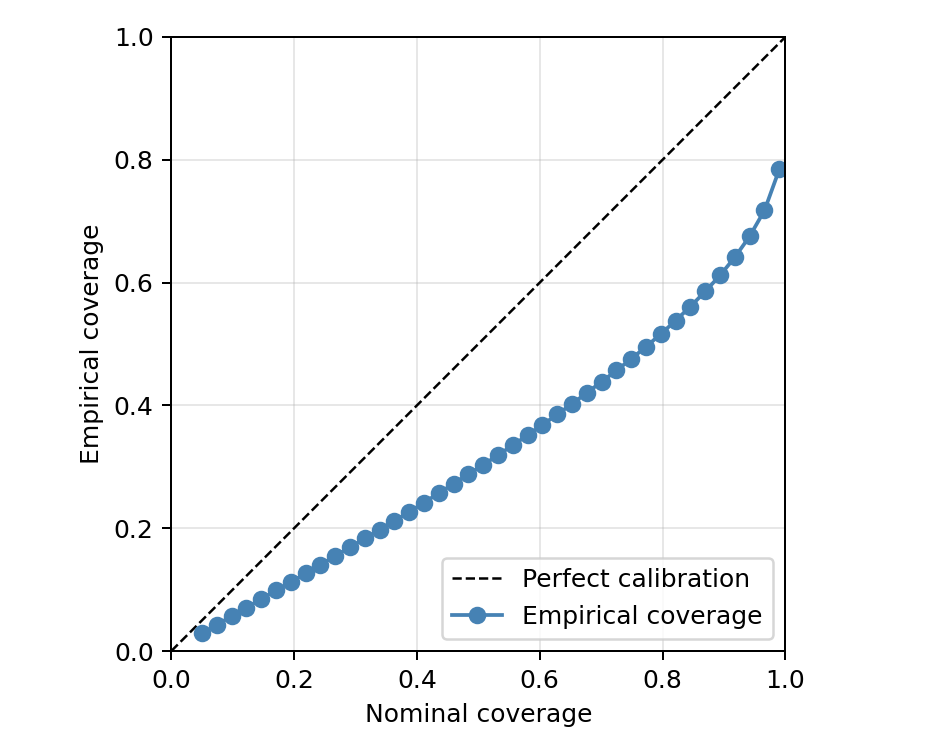}\label{fig:calib-reliability-svh-gp-vae}}
    \\[6pt]
    \subfloat[GP + NMF]{\includegraphics[width=0.24\linewidth]{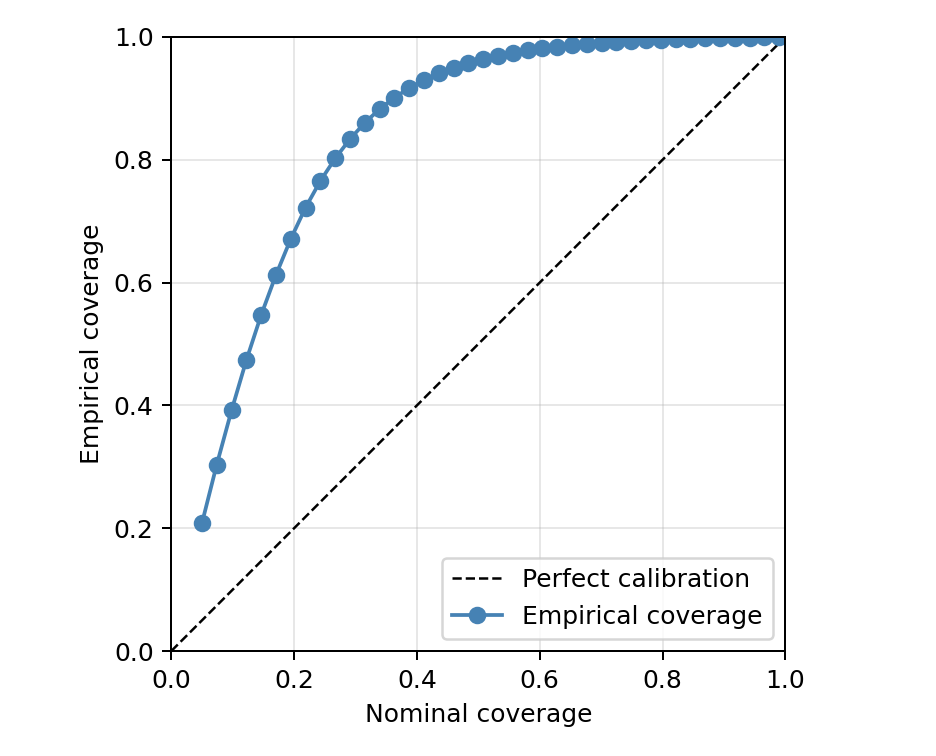}\label{fig:calib-reliability-svh-gp-nmf}}
    \caption{Reliability diagrams on SVH-BD (nominal vs.\ empirical coverage). The dashed diagonal denotes perfect calibration.}
    \label{fig:calib-reliability-svh}
\end{figure}

\begin{figure}[!ht]
    \centering
    \subfloat[1D-CNN]{\includegraphics[width=0.24\linewidth]{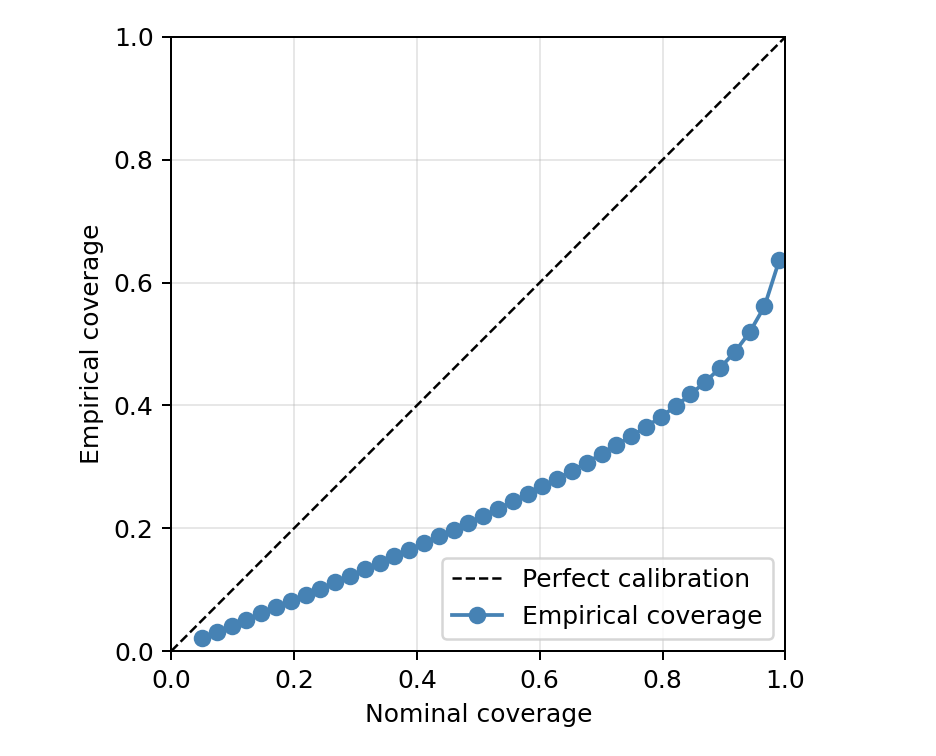}\label{fig:calib-reliability-s3-cnn}}
    \hfill
    \subfloat[P2P]{\includegraphics[width=0.24\linewidth]{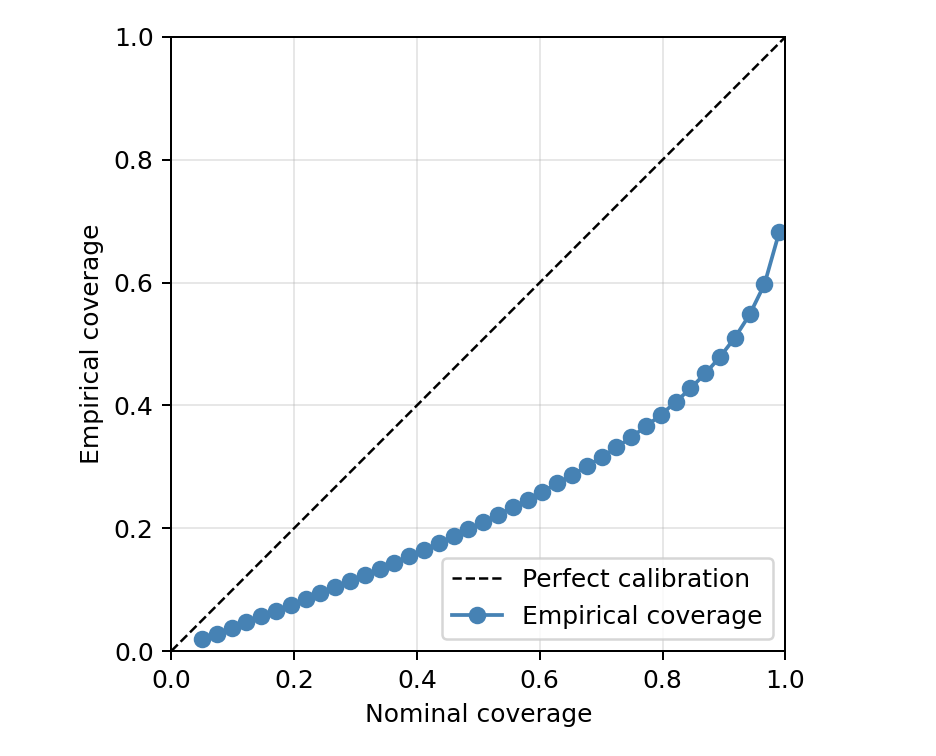}\label{fig:calib-reliability-s3-p2p}}
    \hfill
    \subfloat[FCVAE]{\includegraphics[width=0.24\linewidth]{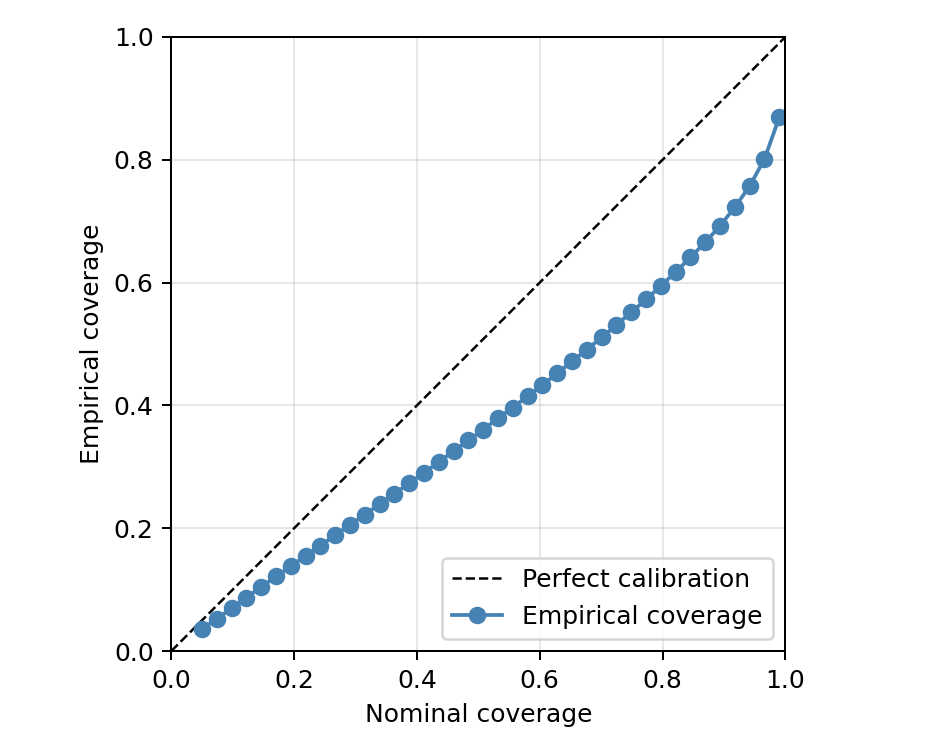}\label{fig:calib-reliability-s3-fcvae}}
    \hfill
    \subfloat[KRR + VAE DEC]{\includegraphics[width=0.24\linewidth]{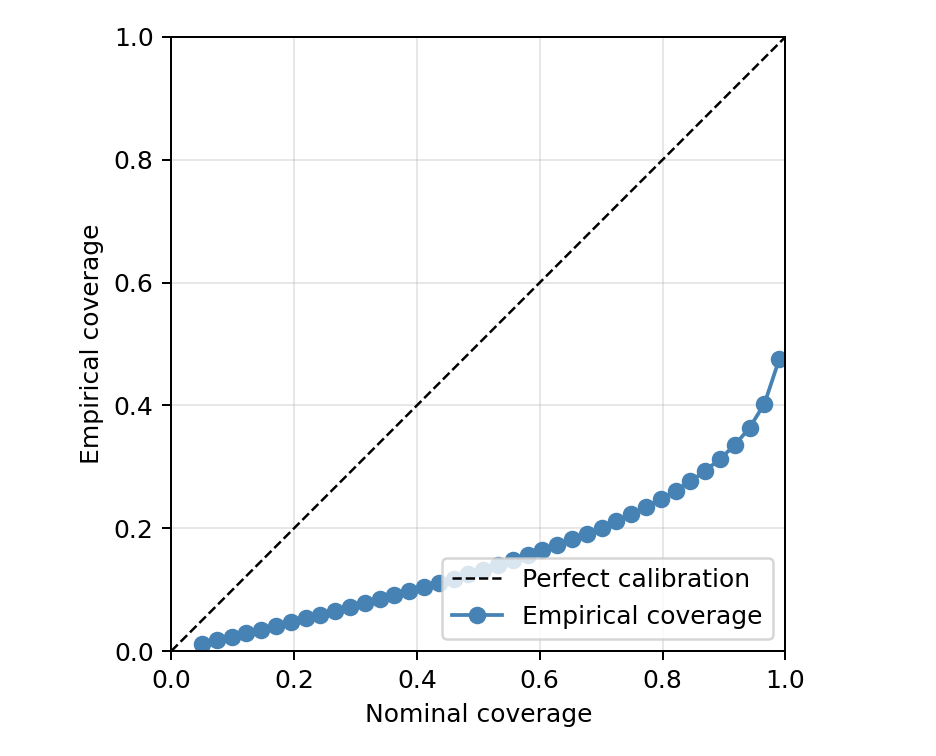}\label{fig:calib-reliability-s3-krr}}
    \caption{Reliability diagrams on Sentinel-3 OLCI (full test patches, no water-only masking). All models are under-calibrated relative to the diagonal, but FCVAE remains closest to perfect calibration.}
    \label{fig:calib-reliability-s3}
\end{figure}

\begin{figure}[!ht]
    \centering
    \subfloat[1D-CNN]{\includegraphics[width=0.48\linewidth]{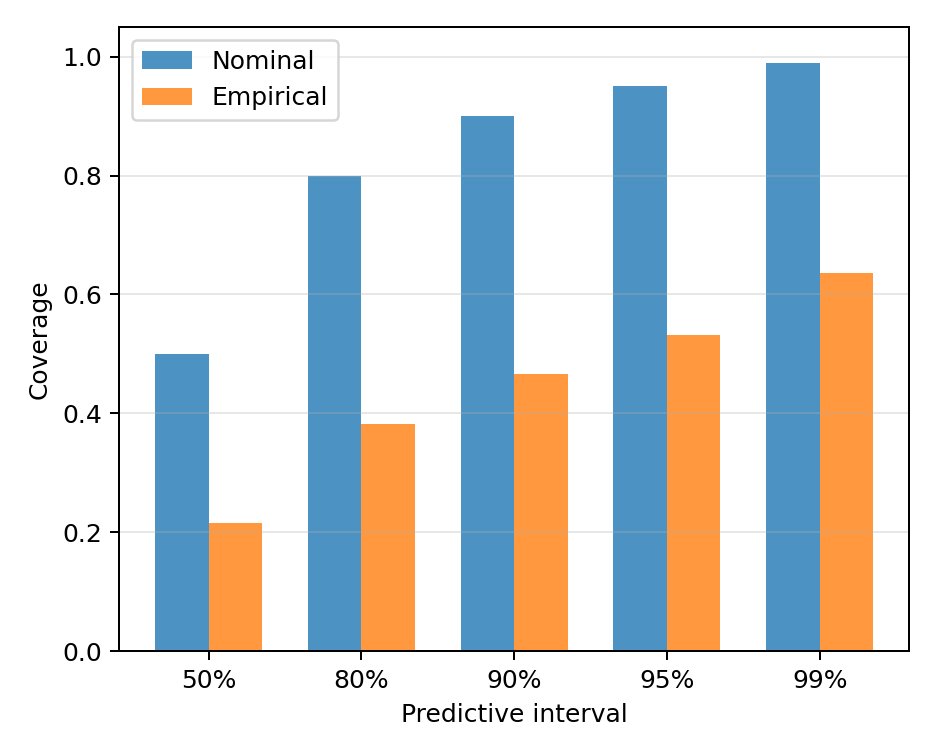}\label{fig:calib-coverage-s3-cnn}}
    \hfill
    \subfloat[FCVAE]{\includegraphics[width=0.48\linewidth]{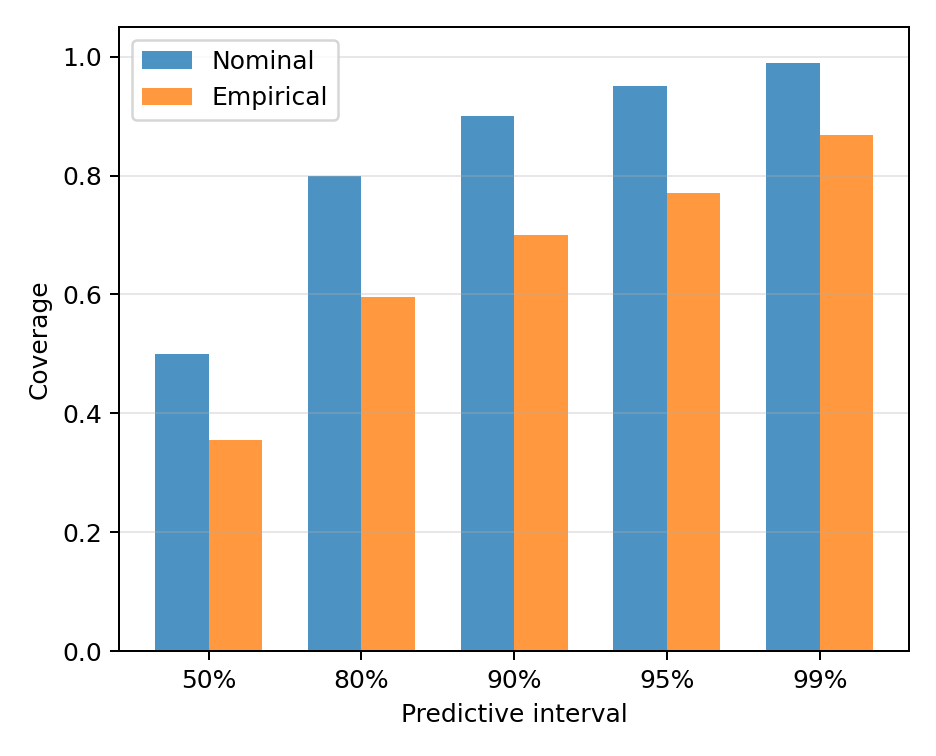}\label{fig:calib-coverage-s3-fcvae}}
    \caption{Nominal vs.\ empirical interval coverage on Sentinel-3 for the most accurate pixel-wise baseline (1D-CNN) and FCVAE. Under-coverage persists across confidence levels for both models, but the gap is smaller for FCVAE (e.g., 70\% empirical vs.\ 90\% nominal at the 90\% level, compared with 47\% for 1D-CNN).}
    \label{fig:calib-coverage-s3}
\end{figure}

%% file: appendix/latent_dimension.tex
\section{Latent-dimension sensitivity}
\label{app:latent-dimension}

\begin{figure}[htbp]
    \centering
    \includegraphics[width=\textwidth]{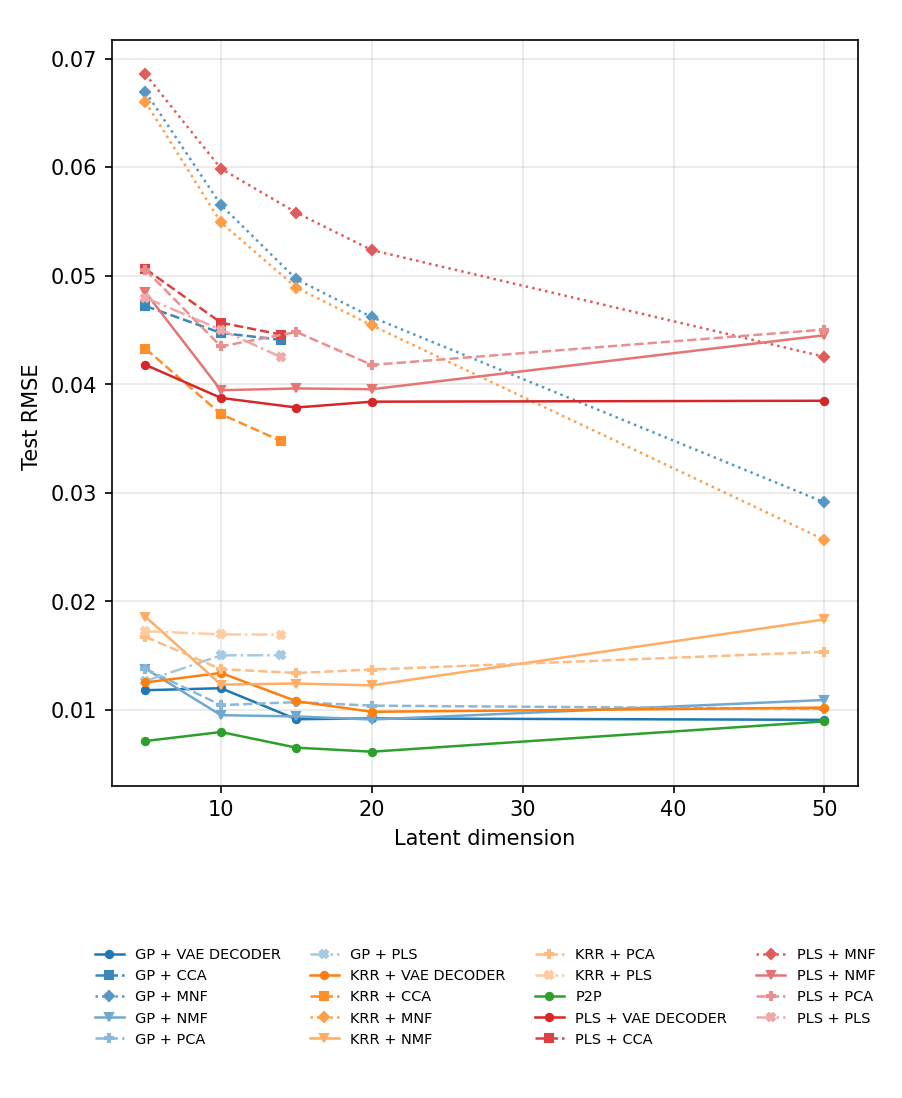}
    \caption{RMSE on the PROSAIL test subset versus latent dimension for the best regressor--DR pairing of each family.}
    \label{fig:dr_size_rmse}
\end{figure}

We evaluate pixel-based regressor--DR combinations on a PROSAIL subset as a function of latent dimension $N$ (FCVAE variants excluded). Figure~\ref{fig:dr_size_rmse} reports RMSE versus $N$. GP + VAE DEC and KRR + VAE DEC achieve the lowest errors and converge quickly; GP paired with NMF or PCA remains competitive at larger $N$. CCA- and PLS-based combinations are weakest, while MNF improves only when $N$ is increased.

